\documentclass{article}

\usepackage{graphicx}
\usepackage{booktabs} 
\usepackage[utf8]{inputenc}
\usepackage{amsmath}
\usepackage{amsfonts}
\usepackage{mathtools}
\usepackage{amssymb}
\usepackage{xcolor}
\usepackage{soul}
\usepackage[noend]{algorithmic}
\usepackage{dsfont}
\usepackage{caption}
\usepackage{subcaption}
\usepackage{todonotes}

\usepackage{hyperref}

 \usepackage[accepted]{mlsys2022}

\begin{document}

\twocolumn[
\mlsystitle{D-Cliques: Compensating for Data Heterogeneity with Topology in
Decentralized Federated Learning}




\begin{mlsysauthorlist}
\mlsysauthor{Aur\'elien Bellet}{inria-lille}
\mlsysauthor{Anne-Marie Kermarrec}{epfl}
\mlsysauthor{Erick Lavoie}{epfl}
\end{mlsysauthorlist}

\mlsysaffiliation{epfl}{EPFL, Lausanne, Switzerland}
\mlsysaffiliation{inria-lille}{Inria, Lille, France}

\mlsyscorrespondingauthor{Erick Lavoie}{erick.lavoie@epfl.ch}

\mlsyskeywords{Decentralized Learning, Federated Learning, Topology,
Heterogeneous Data, Stochastic Gradient Descent}

\vskip 0.3in

\begin{abstract}
The convergence speed of machine learning models trained with Federated
Learning is significantly affected by heterogeneous data partitions, even more
so in a fully decentralized setting without a central server. In this paper, we show that the impact of
label distribution skew, an important type of data heterogeneity, can be
significantly reduced by carefully designing
the underlying communication topology. We present D-Cliques, a novel topology
that reduces gradient bias by grouping nodes in sparsely interconnected
cliques such that the label distribution in a clique is representative
of the global label distribution. We also show how to adapt the updates of
decentralized SGD
to obtain unbiased gradients and implement an effective momentum with
D-Cliques. Our extensive empirical evaluation on MNIST and CIFAR10 demonstrates that our approach
provides similar convergence speed as a fully-connected topology, which provides the best convergence
 in a data heterogeneous setting, with a
significant reduction in the number of edges and messages. In a 1000-node
topology, D-Cliques require 98\% less edges and 96\% less total messages,
with further possible gains using a small-world topology across cliques.
\end{abstract}
]



\printAffiliationsAndNotice{}  


\section{Introduction}

Machine learning is currently shifting from a \emph{centralized}
paradigm, where training data is located on a single
machine or
in a data center, to \emph{decentralized} ones in which data is processed
where it was naturally produced.
This shift is illustrated by the rise of Federated
Learning
(FL)~\cite{mcmahan2016communication}. FL allows
several parties (hospitals, companies, personal
devices...) to collaboratively train machine learning models
on their joint
data without centralizing it. Not only does FL
avoid the costs of moving data, but it also  mitigates privacy and
confidentiality concerns~\cite{kairouz2019advances}.
Yet, working with natural data distributions introduces new challenges for
learning systems, as
local datasets
reflect the usage and production patterns specific to each participant: in
other words, they are
\emph{heterogeneous}. An important type of data heterogeneity encountered in
federated classification problems, known as \emph{label distribution skew} 
\cite{kairouz2019advances,quagmire}, occurs when the frequency of different
classes of examples varies significantly across local datasets.
One of the key challenges in FL is to design algorithms that
can efficiently deal with such heterogeneous data distributions
\cite{kairouz2019advances,fedprox,scaffold,quagmire}.

Federated learning algorithms can be classified into two categories depending
on the underlying network topology they run on. In server-based FL, the
network is organized according to a star topology: a central server orchestrates the training process by
iteratively aggregating model updates received from the participants
(\emph{clients}) and sending back the aggregated model \cite{mcmahan2016communication}. In contrast,
fully decentralized FL algorithms operate over an arbitrary network topology
where participants communicate only with their direct neighbors
in the network. A classic example of such algorithms is Decentralized
SGD (D-SGD) \cite{lian2017d-psgd}, in which participants alternate between
local SGD updates and model averaging with neighboring nodes.

In this paper, we focus on fully decentralized algorithms as they can
generally scale better to the large number of participants seen in ``cross-device''
applications \cite{kairouz2019advances}. Effectively, while a central
server may quickly become a bottleneck as the number of participants increases, the topology used in fully decentralized algorithms can remain sparse
enough such that all participants need only to communicate with a small number of other participants, i.e. nodes have small (constant or logarithmic) degree 
\cite{lian2017d-psgd}. In the homogeneous setting where data is
independent and identically distributed (IID) across nodes, recent work
has shown both empirically
\cite{lian2017d-psgd,Lian2018} and theoretically \cite{neglia2020} that sparse
topologies like rings or grids
do not significantly affect the convergence
speed compared to using denser topologies.



\begin{figure*}[t]
     \centering
     \begin{subfigure}[b]{0.25\textwidth}
         \centering
         \includegraphics[width=\textwidth]{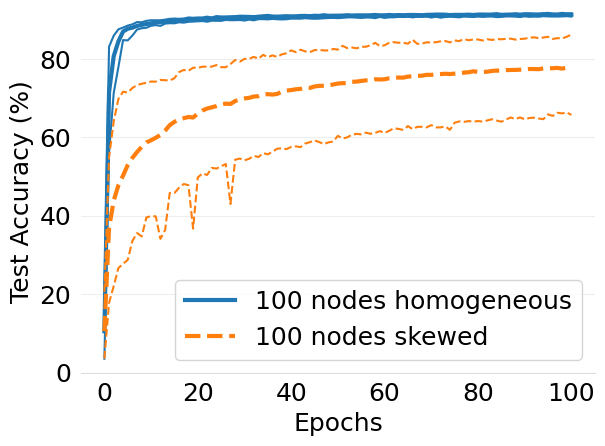}
\caption{\label{fig:ring-IID-vs-non-IID-uneq-classes} Ring topology}
     \end{subfigure}
     \quad
     \begin{subfigure}[b]{0.25\textwidth}
         \centering
         \includegraphics[width=\textwidth]{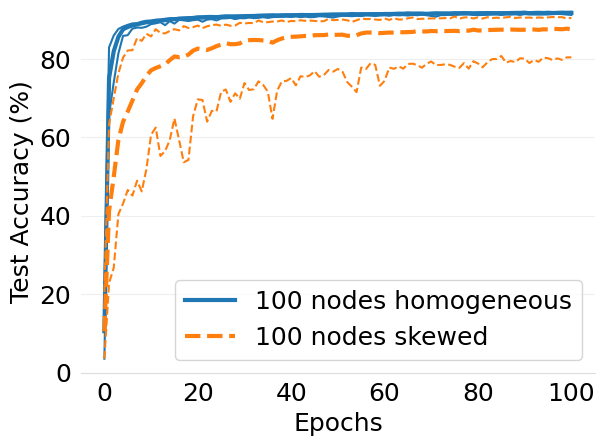}
\caption{\label{fig:grid-IID-vs-non-IID-uneq-classes} Grid topology}
     \end{subfigure}
     \quad
     \begin{subfigure}[b]{0.25\textwidth}
         \centering
         \includegraphics[width=\textwidth]{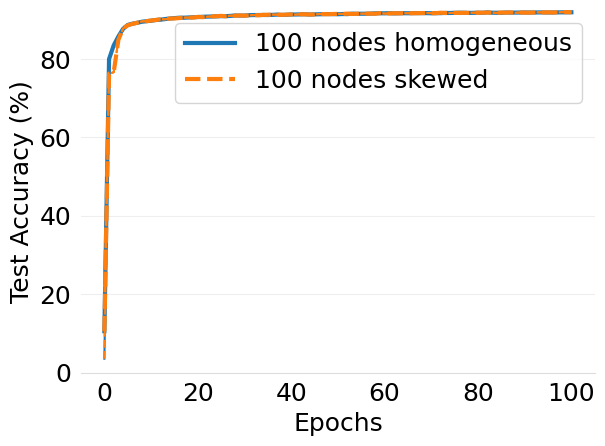}
\caption{\label{fig:fully-connected-IID-vs-non-IID-uneq-classes} Fully-connected topology}
     \end{subfigure}
        \caption{Convergence speed of decentralized
        SGD with and without label distribution skew for different topologies.
        The task is logistic regression on MNIST (see
        Section~\ref{section:experimental-settings} for details on
        the experimental setup). Bold lines show the
        average test
        accuracy across nodes
        while thin lines show the minimum
        and maximum accuracy of individual nodes. While the effect of topology
        is negligible for homogeneous data, it is very significant in the
        heterogeneous case. On a fully-connected network, both cases converge
        similarly.}
        \label{fig:iid-vs-non-iid-problem}
\end{figure*}

In contrast to the homogeneous case however, our experiments demonstrate that 
\emph{the impact of topology is extremely significant for heterogeneous data}.
This phenomenon is illustrated in Figure~\ref{fig:iid-vs-non-iid-problem}: we observe that under
label distribution skew, using a
sparse topology (a ring or
a grid) clearly jeopardizes the convergence speed of decentralized SGD.
We stress the fact
that, unlike in centralized FL
\cite{mcmahan2016communication,scaffold,quagmire}, this
happens even when nodes perform a single local update before averaging the
model with their neighbors. In this paper, we thus address the following
question:

\textit{Can we design sparse topologies with  convergence
  speed similar to a  fully connected network for problems involving
  many participants with label distribution skew?}

Specifically, we make the following contributions:
(1) We propose D-Cliques, a sparse topology in which nodes are organized in
interconnected cliques (i.e., locally fully-connected sets of nodes) such that
the joint label distribution of each clique is close to that of the global 
distribution; (2) We design Greedy Swap, a randomized greedy algorithm for
constructing such cliques efficiently;
 (3) We introduce Clique Averaging, a modified version of 
the standard D-SGD algorithm which decouples gradient averaging, used for
optimizing local models, from distributed averaging, used to ensure that all
models converge, thereby reducing the bias introduced by inter-clique
connections; 
(4) We show how Clique Averaging can be used to implement unbiased momentum
that would otherwise be detrimental in the heterogeneous setting; (5) We 
demonstrate
through an extensive experimental study that our approach  removes the effect
of label distribution skew when training a linear
model and a deep
convolutional network on the MNIST 
and CIFAR10 
datasets respectively;  (6) Finally, we demonstrate the scalability of our
approach by considering  up to 1000-node networks, in contrast to most
previous work on fully decentralized learning which performs empirical
evaluations on networks with
at most a few tens
of nodes
\cite{tang18a,neglia2020,momentum_noniid,cross_gradient,consensus_distance}.

For instance, our results show that under strong label distribution shift,
using D-Cliques in a 1000-node network
requires 98\% less edges ($18.9$ vs $999$ edges per participant on average) to obtain a similar convergence speed as a fully-connected topology,
thereby yielding a 96\% reduction in the total number of required messages 
(37.8 messages per round per node on average instead of 999). Furthermore an additional 22\% improvement
is possible when using a small-world inter-clique topology, with further
potential gains at larger scales through a quasilinear $O(n
\log n)$ scaling in the number of nodes $n$.

The rest of this paper is organized as follows.
We first describe the problem setting in Section~\ref{section:problem}. We
then present the design of D-Cliques in Section~\ref{section:d-cliques}.
Section~\ref{section:evaluation}
compares D-Cliques to different topologies 
and algorithmic variations to demonstrate their benefits, constructed with and without Greedy Swap
in an extensive experimental study. Finally, we review some related work
in Section~\ref{section:related-work}, and conclude with promising directions
for future work in Section~\ref{section:conclusion}.

\section{Problem Setting}

\label{section:problem}

\paragraph{Objective.} We consider a set $N = \{1, \dots, n \}$ of $n$ nodes
seeking to
collaboratively solve a classification task with $L$ classes. We denote a
labeled data point by a tuple $(x,y)$ where $x$ represents the data point 
(e.g., a feature vector) and $y\in\{1,\dots,L\}$ its label.
Each
node has
access to a local dataset that
 follows its own local distribution $D_i$ which may differ from that of other
 nodes.
In this work, we tackle \emph{label distribution skew}: formally, this means
that the
probability of $(x,y)$ under the local distribution $D_i$ of node $i$, denoted
by $p_i(x,y)$,
decomposes as $p_i(x,y)=p(x|y)p_i(y)$, where $p_i(y)$ may vary across nodes.
We
refer to 
\cite{kairouz2019advances,quagmire} for concrete examples of problems
with label distribution skew.

The objective is to find the parameters
$\theta$ of a global model that performs well on the union of the local
 distributions by
 minimizing
 the average training loss:
\begin{equation}
\min_{\theta} \frac{1}{n}\sum_{i=1}^{n} \mathds{E}_
{(x_i,y_i) \sim D_i} [F_i(\theta;x_i,y_i)],
\label{eq:dist-optimization-problem}
\end{equation}
where $(x_i,y_i)$ is a data point drawn from $D_i$ and $F_i$ is the loss
function
on node $i$. Therefore, $\mathds{E}_{(x_i,y_i) \sim D_i} F_i(\theta;x_i,y_i)$
denotes 
the
expected loss of model $\theta$ over $D_i$.

To collaboratively solve Problem \eqref{eq:dist-optimization-problem}, each
node can exchange messages with its neighbors in an undirected network graph
$G=(N,E)$ where $\{i,j\}\in E$ denotes an edge (communication channel)
between nodes $i$ and $j$.

\paragraph{Training algorithm.}
In this work, we use the popular Decentralized Stochastic
Gradient Descent algorithm, aka D-SGD~\cite{lian2017d-psgd}. As
shown in Algorithm~\ref{Algorithm:D-PSGD},
a single iteration of D-SGD at node $i$ consists in sampling a mini-batch
from its local distribution
$D_i$, updating its local model $\theta_i$ by taking a stochastic gradient
descent
(SGD) step according to the mini-batch, and performing a weighted average of
its local model with those of its
neighbors.
This weighted average is defined by a
mixing matrix $W$, in which $W_{ij}$ corresponds to the weight of
the outgoing connection from node $i$ to $j$ and $W_{ij} = 0$ for $
\{i,j\}\notin
E$. To ensure that the local models converge on average to a stationary
point
of Problem
\eqref{eq:dist-optimization-problem}, $W$
must be doubly
stochastic ($\sum_{j \in N} W_{ij} = 1$ and $\sum_{j \in N} W_{ji} = 1$) and
symmetric, i.e. $W_{ij} = W_{ji}$~\cite{lian2017d-psgd}.
Given a network topology $G=(N,E)$, we generate a valid $W$ by computing
standard
Metropolis-Hasting weights~\cite{xiao2004fast}:
\begin{equation}
  W_{ij} = \begin{cases}
    \frac{1}{\max(\text{degree}(i), \text{degree}(j)) + 1} & \text{if}~i \neq
    j \text{ and } \{i,j\}\in E,\\
   1 - \sum_{j \neq i} W_{ij} & \text{if } i = j, \\
   0 & \text{otherwise}.
  \end{cases}
  \label{eq:metro}
\end{equation}

\begin{algorithm}[t]
   \caption{D-SGD, Node $i$}
   \label{Algorithm:D-PSGD}
   \begin{algorithmic}[1]
        \STATE \textbf{Require:} initial model $\theta_i^{(0)}$,
        learning rate $\gamma$, mixing weights $W$, mini-batch size $m$,
        number of steps $K$
        \FOR{$k = 1,\ldots, K$}
          \STATE $S_i^{(k)} \gets \text{mini-batch of $m$ samples drawn
          from~} D_i$
          \STATE $\theta_i^{(k-\frac{1}{2})} \gets \theta_i^{(k-1)} - \gamma
          \nabla F(\theta_i^{(k-1)}; S_i^{(k)})$ 
          \STATE $\theta_i^{(k)} \gets \sum_{j \in N} W_{ji}^{(k)} \theta_j^{(k-\frac{1}{2})}$
        \ENDFOR
   \end{algorithmic}
\end{algorithm}

\section{D-Cliques}
\label{section:d-cliques}

In this section, we introduce D-Cliques, a topology
designed to compensate for data heterogeneity. We also present some
modifications of D-SGD that leverage some properties of the proposed
topology and allow to implement a successful momentum scheme.

\subsection{Intuition}

\begin{figure}[t]
     \centering
     \begin{subfigure}[b]{0.18\textwidth}
         \centering
         \includegraphics[width=\textwidth]{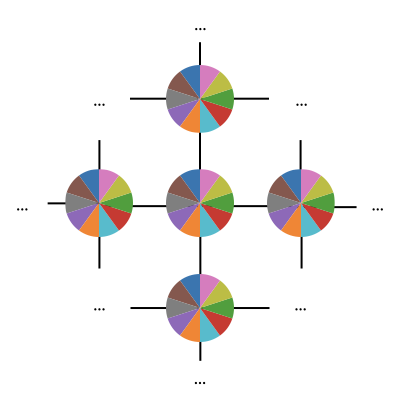}
\caption{\label{fig:grid-iid-neighbourhood} Homogeneous data}
     \end{subfigure}
     \hspace*{.5cm}
     \begin{subfigure}[b]{0.18\textwidth}
         \centering
         \includegraphics[width=\textwidth]{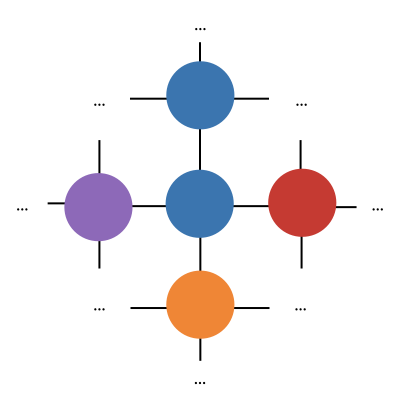}
\caption{\label{fig:grid-non-iid-neighbourhood}  Heterogeneous data}
     \end{subfigure}
        \caption{Neighborhood in a grid.}
        \label{fig:grid-iid-vs-non-iid-neighbourhood}
\end{figure}

To give the intuition behind
our approach, let us consider the neighborhood of a single node in a grid
topology represented
on Figure~\ref{fig:grid-iid-vs-non-iid-neighbourhood}.
Nodes are distributed randomly in the grid and the colors of a node represent
the proportion of each class in its local dataset. In the homogeneous
setting, the label distribution is the same across
nodes: in the example shown in Figure~\ref{fig:grid-iid-neighbourhood}, all classes
are represented in equal proportions on all nodes. This is not the case in the
heterogeneous setting: Figure~\ref{fig:grid-non-iid-neighbourhood} shows an
extreme case of label distribution skew where each
node holds examples of a single class only.

From the point of view of the center node in
Figure~\ref{fig:grid-iid-vs-non-iid-neighbourhood}, a single training step of
D-SGD is
equivalent to sampling a mini-batch five times larger from the union of the
local distributions of neighboring nodes.
In the homogeneous case, since gradients are computed from examples of all
classes,
the resulting averaged gradient  points in a direction that tends to reduce
the loss across all classes. In contrast, in the heterogeneous case, the
representation of classes in the immediate neighborhood of the node is
different from the global label distribution
(in Figure~\ref{fig:grid-non-iid-neighbourhood}, only a
subset of classes are represented), thus the gradients will
be biased.
Importantly, as the distributed averaging process takes several steps to
converge, this variance persists across iterations as the locally computed
gradients are far from the global average.\footnote{One could perform a
sufficiently large number of
averaging steps between each gradient step, but this is too costly in
practice.} This can significantly slow down
convergence speed to the point of making decentralized optimization
impractical.

With D-Cliques, we address label distribution skew by
carefully designing a
network topology composed of \textit{locally representative cliques} while 
maintaining \textit{sparse inter-clique connections} only.

\subsection{Constructing Locally Representative Cliques}

D-Cliques construct a topology in which each node is part of a \emph{clique} 
(i.e., a subset of nodes whose induced subgraph is fully connected)
such that the label distribution in each clique is
close to the global label distribution. Formally, for a label $y$ and a
clique composed of nodes $C\subseteq N$, we denote by $p_C(y)=
\frac{1}{|C|}\sum_{i\in C} p_i(y)$ the distribution of $y$ in $C$
and by $p(y)=\frac{1}{n}\sum_{i\in N} p_i(y)$ its global distribution.
We measure the \textit{skew} of $C$ by the sum
of the absolute differences of $p_C(y)$ and $p(y)$:
\begin{equation}
\label{eq:skew}
        \textit{skew}(C) =\
    \sum_{l=1}^L | p_C(y = l) - p(y = l) |.
\end{equation}


To efficiently construct a set of cliques with small skew, we propose
Greedy-Swap (Algorithm~\ref{Algorithm:greedy-swap}). The parameter
$M$ is the maximum size of cliques and controls the
number of intra-clique edges. We start by initializing cliques at
random. Then, for
a certain number of steps $K$, we randomly pick two cliques and swap two of
their nodes so as to decrease the sum of skews of the two cliques. The swap is
chosen randomly among the ones that decrease the skew, hence
this algorithm can be seen as a form of randomized greedy algorithm.
We note that this algorithm only requires
the knowledge of the label distribution $p_i(y)$ at each node $i$. For the
sake of
simplicity, we assume that D-Cliques are constructed from the global
knowledge of these distributions, which can easily be obtained by
decentralized averaging in a pre-processing step \citep[e.g.,][]
{jelasity2005largegossip}.

\begin{algorithm}[t]
   \caption{D-Cliques Construction via Greedy Swap}
   \label{Algorithm:greedy-swap}
   \begin{algorithmic}[1]
        \STATE \textbf{Require:} maximum clique size $M$, max steps $K$, set
        of all nodes $N = \{ 1, 2, \dots, n \}$,
        procedure $\texttt{inter}(\cdot)$ to create intra-clique connections
        (see Sec.~\ref{section:interclique-topologies}) 
         \STATE $DC \leftarrow []$ 
         \WHILE {$N \neq \emptyset$}
         \STATE $C \leftarrow$ sample $M$ nodes from $N$ at random
         \STATE $N \leftarrow N \setminus C$; $DC.\text{append}(C)$
         \ENDWHILE
         \FOR{$k \in \{1, \dots, K\}$}
        \STATE $C_1,C_2 \leftarrow$ random sample of 2 elements from $DC$
          \STATE $s \leftarrow \textit{skew}(C_1) + skew(C_2)$
        \STATE $\textit{swaps} \leftarrow []$
        \FOR{$i \in C_1, j \in C_2$}
          \STATE $s' \leftarrow \textit{skew}(C_1\setminus\{i\}\cup\{j\})
          + \textit{skew}(C_2 \setminus\{i\}\cup\{j\})$\hspace*{-.05cm}
          \IF {$s' < s$}
            \STATE \textit{swaps}.append($(i, j)$)
          \ENDIF
        \ENDFOR
        \IF {len(\textit{swaps}) $> 0$}
          \STATE $(i,j) \leftarrow$ random element from $
          \textit{swaps}$ 
          \STATE $C_1 \leftarrow C_1 \setminus\{i\}\cup\{j\}; C_2 \leftarrow C_2 \setminus\{j\}\cup\{i\}$
        \ENDIF
         \ENDFOR
        \STATE $E\leftarrow \{(i,j) : C\in DC, i,j\in C, i\neq j\}$
        \RETURN topology $G=(N,E \cup 
        \texttt{inter}(DC))$
   \end{algorithmic}
\end{algorithm}

The key idea of D-Cliques is to ensure the clique-level label distribution
$p_C(y)$
 matches closely the global distribution $p(y)$. As a consequence,
the local models of nodes across cliques remain rather close. Therefore, a
sparse inter-clique topology can be used, significantly reducing the total
number of edges without slowing down the convergence. We discuss some possible
choices for this inter-clique topology in the next section.

\subsection{Adding Sparse Inter-Clique Connections}
\label{section:interclique-topologies}

To ensure a global consensus and convergence, we introduce
\textit{inter-clique connections} between a small number of node pairs that
belong to different cliques, thereby implementing the \texttt{inter}
procedure called at the end of Algorithm~\ref{Algorithm:greedy-swap}.
We aim to ensure that the degree of each node remains low and balanced so as
to make the network topology well-suited to decentralized federated learning.
We consider several choices of inter-clique topology, which offer
different scalings for the number of required edges and the average distance
between nodes in the resulting graph.

The \textit{ring} has (almost) the fewest possible number of edges for the
graph to be connected: in this case, each clique is connected to exactly
two other cliques by a single edge. This topology requires only $O(\frac{n}
{M})$ inter-clique edges but suffers an $O(n)$ average distance between nodes.

The
\textit{fractal} topology
provides a logarithmic bound on the average distance. In this
hierarchical scheme, cliques are arranged in larger groups of $M$ cliques that
are connected
internally with one edge per
pair of cliques, but with only one edge between pairs of larger groups. The
topology is built recursively such that $M$ groups will themselves form a
larger group at the next level up. This results in at most $M$ edges per node 
if edges are evenly distributed: i.e., each group within the same level adds 
at most $M-1$ edges to other groups, leaving one node per group with $M-1$ 
edges that can receive an additional edge to connect with other groups at the next level.
Since nodes have at most $M$ edges, the total number of inter-clique edges
is at most $nM$ edges.

We can also design an inter-clique topology in which the number of edges
scales in a log-linear fashion by following a
small-world-like topology~\cite{watts2000small} applied on top of a
ring~\cite{stoica2003chord}. In this scheme, cliques are first arranged in a
ring. Then each clique adds symmetric edges, both clockwise and
counter-clockwise on the ring, with the $c$ closest cliques in sets of
cliques that are exponentially bigger the further they are on the ring (see
Algorithm~\ref{Algorithm:Smallworld} in Appendix~\ref{app:small_world} for
details on the construction). This topology ensures a good connectivity with
other cliques that are close on the ring, while keeping the average
distance small. This scheme uses $O(c\frac{n}{M}\log\frac{n}{M})$ edges,
i.e.
log-linear in $n$.

\begin{figure}[t]
    \centering
    \includegraphics[width=0.20\textwidth]{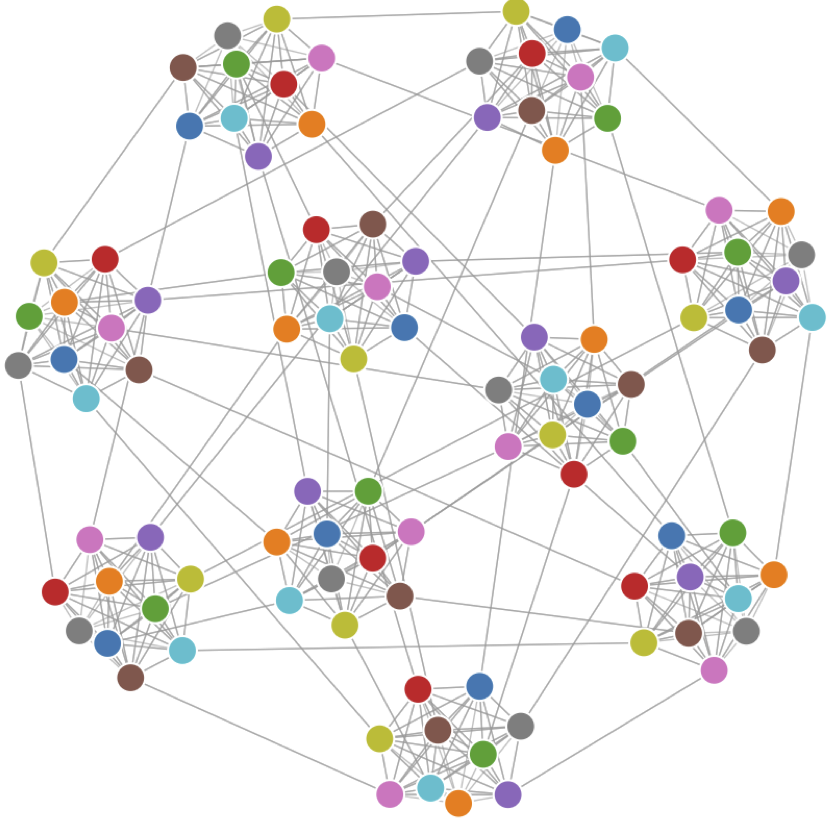}
    \caption{\label{fig:d-cliques-figure} D-Cliques with $n=100$, $M=10$ and a
fully connected inter-clique topology on a problem with 1 class/node.}
\end{figure}

Finally, we can consider a \emph{fully connected} inter-clique topology
 such that each clique has exactly
one edge with each of the other cliques, spreading these additional edges
equally among the nodes of a clique, as illustrated in Figure~\ref{fig:d-cliques-figure}. 
This has the advantage of
bounding the distance between any pair of nodes to $3$ but requires
$O(\frac{n^2}{M^2})$ inter-clique edges, i.e. quadratic in $n$.

\subsection{Optimizing over D-Cliques with Clique Averaging and Momentum}
\label{section:clique-averaging-momentum}



\begin{figure}[t]
         \centering
         \includegraphics[width=0.3\textwidth]{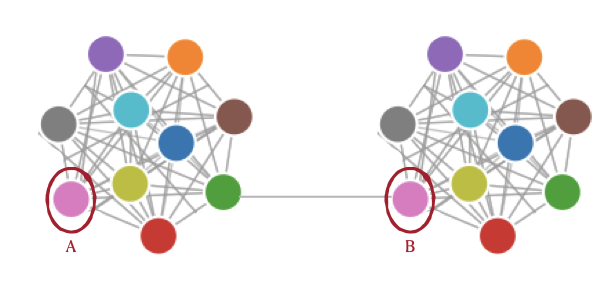}
\caption{\label{fig:connected-cliques-bias} Illustrating the bias induced by
inter-clique connections (see main text for details).}
\end{figure}

While limiting the number of inter-clique connections reduces the
amount of messages traveling on the network, it also introduces a form of
bias.
Figure~\ref{fig:connected-cliques-bias} illustrates the problem on the
simple case of two cliques connected by one inter-clique edge (here,
between the green node of the left clique and the pink node of the right
clique). In this example, each node holds example of a single class. Let us
focus on node A. With weights computed as in \eqref{eq:metro},
node A's self-weight is $\frac{12}
{110}$, the weight between A and the green node connected to B is
$\frac{10}{110}$, and
all other neighbors of A have a weight of $\frac{11}{110}$. Therefore, the
gradient at A is biased towards its own class (pink) and against the green
class. A similar bias holds for all other nodes
without inter-clique edges with respect to their respective classes. For node
B, all its edge weights (including its self-weight) are equal to $\frac{1}
{11}$. However, the green class is represented twice (once as a clique
neighbor and once from the inter-clique edge), while all other classes are
represented only once. This biases the gradient toward the green class. The
combined effect of these two sources of bias is to increase the variance
of the local models across nodes.

\paragraph{Clique Averaging.} 
We address this problem by adding \emph{Clique
Averaging} to D-SGD
(Algorithm~\ref{Algorithm:Clique-Unbiased-D-PSGD}), which essentially
decouples gradient averaging from model averaging. The idea is to use only the
gradients of neighbors within the same clique to compute the average gradient
so as to remove the bias due to inter-clique edges. In contrast, all
neighbors' models (including those in different cliques)
participate in model averaging as in the original version. Adding Clique Averaging
requires gradients to be sent separately from the model parameters: the number
of messages
exchanged between nodes is therefore twice their number of edges.

\begin{algorithm}[t]
   \caption{D-SGD with Clique Averaging, Node $i$}
   \label{Algorithm:Clique-Unbiased-D-PSGD}
   \begin{algorithmic}[1]
        \STATE \textbf{Require} initial model $\theta_i^{(0)}$, learning
        rate $\gamma$, mixing weights $W$, mini-batch size $m$, number of
        steps $K$
        \FOR{$k = 1,\ldots, K$}
          \STATE $S_i^{(k)} \gets \text{mini-batch of $m$ samples drawn
          from~} D_i$
          \STATE $g_i^{(k)} \gets \frac{1}{|\textit{Clique}(i)|}\sum_{j \in 
          \textit{Clique(i)}}  \nabla F(\theta_j^{(k-1)}; S_j^{(k)})$
          \STATE $\theta_i^{(k-\frac{1}{2})} \gets \theta_i^{(k-1)} - \gamma g_i^{(k)}$ 
          \STATE $\theta_i^{(k)} \gets \sum_{j \in N} W_{ji}^{(k)} \theta_j^{(k-\frac{1}{2})}$
        \ENDFOR
   \end{algorithmic}
\end{algorithm}

\paragraph{Implementing momentum with Clique Averaging.}
Efficiently training high capacity models usually requires additional
optimization techniques. In particular, momentum~\cite{pmlr-v28-sutskever13}
increases the magnitude of the components of the gradient that are shared
between several consecutive steps, and is critical for deep convolutional networks like
LeNet~\cite{lecun1998gradient,quagmire} to converge quickly. However, a direct
application of momentum in data heterogeneous settings can
actually be very detrimental and even fail to converge, as we will show in
 our experiments (Figure~\ref{fig:cifar10-c-avg-momentum} in
 Section~\ref{section:evaluation}).
Clique Averaging allows us to reduce the bias in the momentum by using the
clique-level average gradient $g_i^{(k)}$ of
Algorithm~\ref{Algorithm:Clique-Unbiased-D-PSGD}:
\begin{equation}
v_i^{(k)} \leftarrow m v_i^{(k-1)} +  g_i^{(k)}.
\end{equation}
It then suffices to modify the original gradient step to apply momentum:
\begin{equation}
\theta_i^{(k-\frac{1}{2})} \leftarrow \theta_i^{(k-1)} - \gamma v_i^{(k)}.
\end{equation}



\section{Evaluation}
\label{section:evaluation}

In this section, we first compare D-Cliques to alternative topologies to
show the benefits and relevance of our main design choices. Then, 
we evaluate different inter-clique topologies to further reduce the number of
inter-clique connections so as to gracefully scale with the number of
nodes. Then, we show the impact of removing intra-clique edges.
 Finally, we show that Greedy Swap
(Alg.~\ref{Algorithm:greedy-swap}) 
constructs cliques efficiently with consistently lower skew than
random cliques.

\subsection{Experimental Setup}
\label{section:experimental-settings}

Our main goal is to provide a fair comparison of the convergence speed across
different topologies and algorithmic variations, in order to
show that D-Cliques
can remove much of the effects of label distribution skew.

We experiment with two datasets: MNIST~\cite{mnistWebsite} and
CIFAR10~\cite{krizhevsky2009learning}, which both have $L=10$ classes.
For MNIST,  we use 50k and 10k examples from the original 60k training 
set for training and validation respectively. We use all 10k examples of 
the test set to measure prediction accuracy.  The validation set preserves the
original unbalanced ratio of the classes in the test set, and the remaining
examples become the training set.
For CIFAR10, classes are evenly balanced: we initially used 45k/50k images 
of the original training set for training, 5k/50k for validation, and all 10k examples 
of the test set for measuring prediction accuracy. After tuning hyper-parameters
on initial experiments, we then used all 50k images of the original training set
for training for all experiments, as the 45k did not split evenly in 1000 nodes
with the partitioning scheme explained in the next paragraph.

For both MNIST and CIFAR10, we use the heterogeneous data partitioning scheme
proposed by~\citet{mcmahan2016communication} 
in their seminal FL work: 
we sort all training examples by class, then split the list into shards of
equal size, and randomly assign two shards to each node. When the number of
examples of one class does not divide evenly in shards, as is the case for MNIST, some shards may have examples of more than one class and therefore nodes may have examples
of up to 4 classes. However, most nodes will have examples of 2 classes.  The varying number 
of classes, as well as the varying distribution of examples within a single node, makes the task 
of creating cliques with low skew nontrivial.

We
use a logistic regression classifier for MNIST, which
provides up to 92.5\% accuracy in the centralized setting.
For CIFAR10, we use a Group-Normalized variant of LeNet~\cite{quagmire}, a
deep convolutional network which achieves an accuracy of $74.15\%$ in the
centralized setting.
These models are thus reasonably accurate (which is sufficient to
study the effect of the topology) while being sufficiently fast to train in a
fully decentralized setting and simple enough to configure and analyze.
Regarding hyper-parameters, we jointly optimize the learning rate and
mini-batch size on the
validation set for 100 nodes, obtaining respectively $0.1$ and $128$ for
MNIST and $0.002$ and $20$ for CIFAR10.
For CIFAR10, we additionally use a momentum of $0.9$.

We evaluate 100- and 1000-node networks by creating multiple models 
in memory and simulating the exchange of messages between nodes.
To ignore the impact of distributed execution strategies and system
optimization techniques, we report the test accuracy of all nodes (min, max,
average) as a function of the number of times each example of the dataset has
been sampled by a node, i.e. an \textit{epoch}. This is equivalent to the classic 
case of a single node sampling the full distribution.
To further make results comparable across different number of nodes, we lower
the batch size proportionally to the number of nodes added, and inversely,
e.g. on MNIST, 128 with 100 nodes vs. 13 with 1000 nodes. This
ensures the same number of model updates and averaging per epoch, which is
important to have a fair comparison.\footnote{Updating and averaging models
after every example can eliminate the impact of label distribution skew. However, the
resulting communication overhead is impractical.}

Finally, we compare our results against an ideal baseline:
a fully-connected network topology with the same number of nodes. 
This baseline is essentially equivalent to a centralized (single) IID node using a batch size
$n$ times bigger, where $n$ is the number of nodes.  Both a fully-connected network and a single IID node
 effectively optimize a single model and sample
uniformly from the global distribution: both therefore remove entirely the
effect of label distribution skew and of the network topology on the
optimization. In practice, we prefer a
fully-connected network because it
 converges slightly faster and obtains slightly 
better final accuracy than a single node sampling randomly from the global
distribution.\footnote{We 
conjecture that an heterogeneous data partition in a fully-connected network may force 
more balanced representation of all classes in the union of all mini-batches, leading to better convergence.}

\subsection{D-Cliques Match the Convergence Speed of Fully-Connected with a
Fraction of the Edges}
\label{section:d-cliques-vs-fully-connected}

\begin{figure}[htbp]
    \centering        
    \begin{subfigure}[b]{0.23\textwidth}
    \centering
    \includegraphics[width=\textwidth]{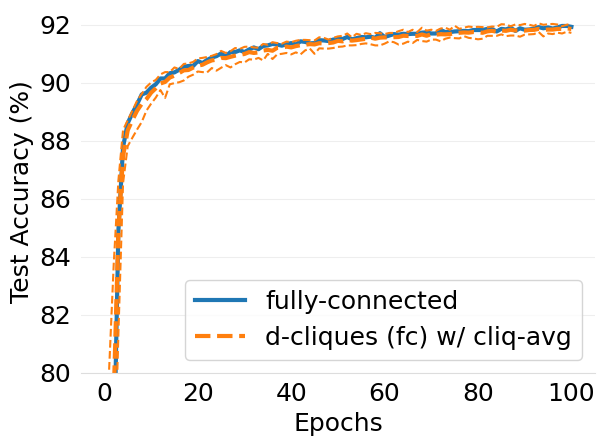}
    \caption{\label{fig:convergence-speed-mnist-dc-fc-vs-fc-2-shards-per-node} MNIST}
    \end{subfigure}
    \hfill
    \begin{subfigure}[b]{0.23\textwidth}
    \centering
    \includegraphics[width=\textwidth]{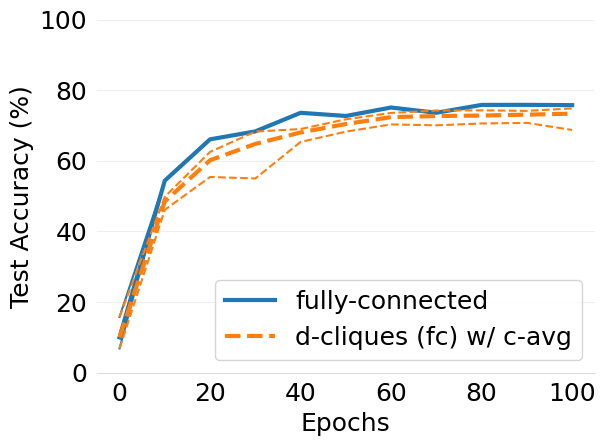}
    \caption{
    \label{fig:convergence-speed-cifar10-dc-fc-vs-fc-2-shards-per-node} CIFAR10 (w/ momentum)}
    \end{subfigure}
\caption{\label{fig:convergence-speed-dc-vs-fc-2-shards-per-node} Comparison on 100 heterogeneous nodes (2 shards/node)
between a fully-connected network and D-Cliques (fully-connected) constructed with Greedy Swap (10 cliques of 10 nodes) using
Clique Averaging. Bold line is the average accuracy over
all nodes. Thinner upper and lower lines are maximum and minimum accuracy over
all nodes.}
\end{figure}

In this first experiment, we show that D-Cliques with Clique Averaging (and
momentum when mentioned) converges 
almost as fast as a fully-connected network on both MNIST and CIFAR10. Figure~\ref{fig:convergence-speed-dc-vs-fc-2-shards-per-node} 
illustrates the convergence speed of D-Cliques with $n=100$ nodes on MNIST (with Clique Averaging) 
and CIFAR10 (with Clique Averaging and momentum). Observe that the convergence speed is
very close to that of a fully-connected topology, and significantly better than with
a ring or a grid (see Figure~\ref{fig:iid-vs-non-iid-problem}). 
It also has less variance than both the ring and grid. 


\subsection{Clique Averaging is Beneficial and Sometimes Necessary}
\label{sec:exp:clique_avg}

\begin{figure}[htbp]
         \centering
         \includegraphics[width=0.23\textwidth]{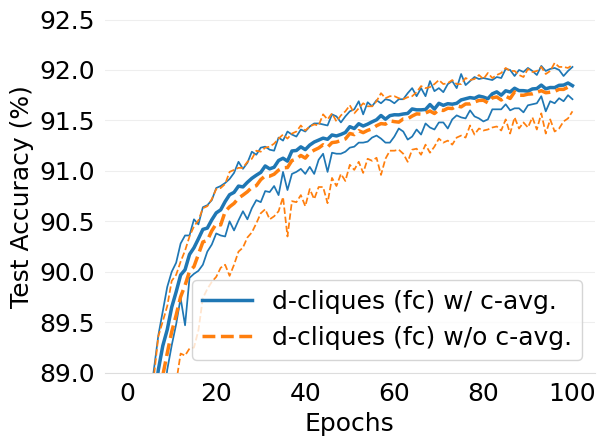}
\caption{\label{fig:d-clique-mnist-clique-avg} MNIST: Effect of Clique Averaging on D-Cliques (fully-connected) with 10 cliques of 10 heterogeneous nodes (100 nodes). Y axis starts at 89.}
\end{figure}

In this experiment, we perform an ablation study of the effect of Clique Averaging.
Figure~\ref{fig:d-clique-mnist-clique-avg} shows that Clique Averaging
(Algorithm~\autoref{Algorithm:Clique-Unbiased-D-PSGD})
 reduces the variance of models across nodes and slightly accelerates the
convergence on MNIST. Recall that Clique Averaging induces a small
additional cost, as gradients
and models need to be sent in two separate rounds of messages. 
Nonetheless, compared to fully connecting all nodes, the total number 
of messages per round for 100 nodes is reduced by $\approx 80\%$.

\begin{figure}[htbp]
    \centering        
    \begin{subfigure}[b]{0.23\textwidth}
    \centering
    \includegraphics[width=\textwidth]{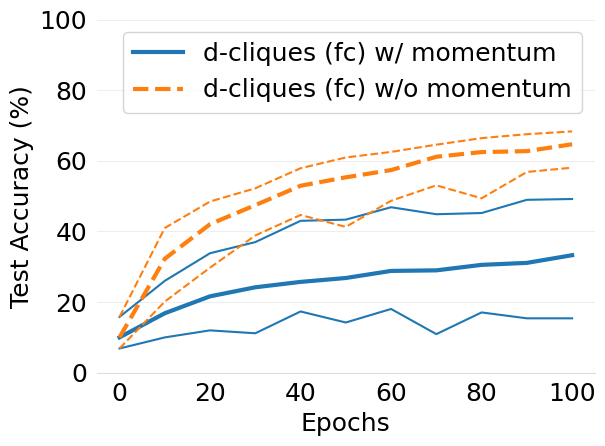}
    \caption{\label{fig:convergence-speed-cifar10-wo-c-avg-no-mom-vs-mom-2-shards-per-node} Without Clique Averaging }
    \end{subfigure}
    \hfill
    \begin{subfigure}[b]{0.23\textwidth}
    \centering
    \includegraphics[width=\textwidth]{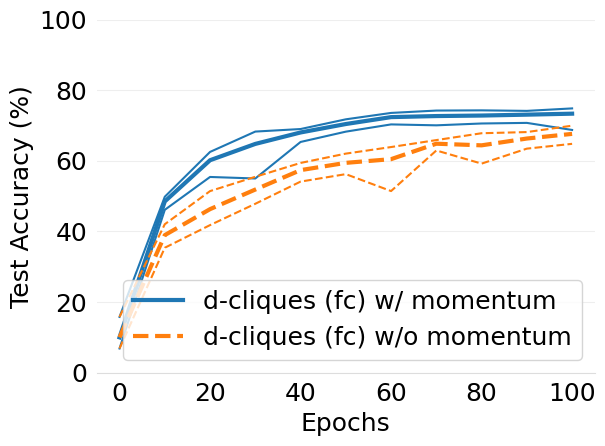}
    \caption{\label{fig:convergence-speed-cifar10-w-c-avg-no-mom-vs-mom-2-shards-per-node} With Clique Averaging}
    \end{subfigure}
\caption{\label{fig:cifar10-c-avg-momentum} CIFAR10: Effect of Clique Averaging, without and with
momentum, on D-Cliques (fully-connected) with 10 cliques of 10 heterogeneous nodes (100 nodes).}
\end{figure}

The effect of Clique Averaging is much more pronounced on CIFAR10, as can
be seen in
Figure~\ref{fig:cifar10-c-avg-momentum}, especially when used in combination with momentum.
Without Clique Averaging,
the use of momentum is actually detrimental. With Clique Averaging, the 
situation reverses and momentum is again beneficial. The combination
of both has the fastest convergence speed and the lowest variance among all
four possibilities. We believe that the gains obtained with Clique
Averaging are larger on CIFAR10 than on MNIST because the model we train on
CIFAR10 (a deep convolutional network) has much higher capacity than the
linear model used for MNIST. The resulting highly nonconvex objective increases the
sensitivity of local updates to small differences in the gradients, making
them point in different directions, as observed by \citet{consensus_distance}
even in the homogeneous setting.
Clique Averaging helps to reduce this effect by reducing the bias in
local gradients.


\subsection{D-Cliques Converge Faster than Random Graphs}
\label{section:d-cliques-vs-random-graphs}

\begin{figure}[htbp]
     \centering     
         \begin{subfigure}[b]{0.23\textwidth}
         \centering
         \includegraphics[width=\textwidth]{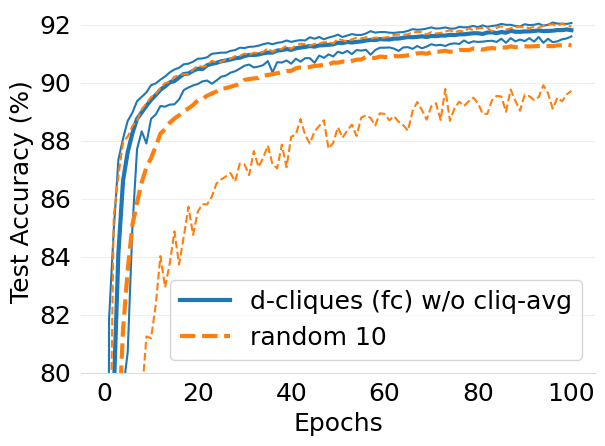}
                  \caption{MNIST}
         \end{subfigure}
                 \hfill                      
        \begin{subfigure}[b]{0.23\textwidth}
        \centering
         \includegraphics[width=\textwidth]{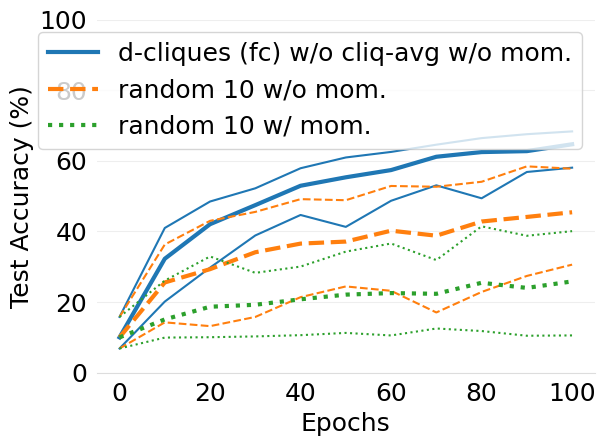}
         \caption{CIFAR10}
     \end{subfigure} 
 \caption{\label{fig:convergence-random-vs-d-cliques-2-shards} Comparison on 100 heterogeneous nodes between D-Cliques (fully-connected) with 10 cliques of size 10 and a random graph with 10 edges per node \textit{without} Clique Averaging or momentum.} 
\end{figure}

In this experiment, we compare D-Cliques to a random graph that has a similar 
number of edges (10) per node to determine
whether a simple sparse topology could work equally well. 
To ensure a fair comparison, because a random graph does not support 
Clique Averaging, we do not use it for D-Cliques either.
\autoref{fig:convergence-random-vs-d-cliques-2-shards} 
shows that even \textit{without} Clique Averaging, D-Cliques converge faster and with
lower variance. Furthermore, the use of momentum in a random graph
is detrimental, similar to D-Cliques without the use of Clique Averaging 
(see \autoref{fig:convergence-speed-cifar10-wo-c-avg-no-mom-vs-mom-2-shards-per-node}).
This shows that a careful design of the topology is indeed necessary.

D-Cliques converge faster even if we were to create diverse neighborhoods 
in a random graph with lower skew and used those to unbias gradients in an analogous 
way to Clique Averaging (details in Annex~\ref{section:d-cliques-clustering-is-necessary}, as 
the experiments require a different partitioning scheme for a fair comparison).
The clustering provided by D-Cliques therefore provides faster convergence.



\subsection{D-Cliques Scale with Sparser Inter-Clique Topologies}
\label{section:scaling}

\begin{figure}[htbp]
     \centering
     \begin{subfigure}[b]{0.23\textwidth}
         \centering
            \includegraphics[width=\textwidth]{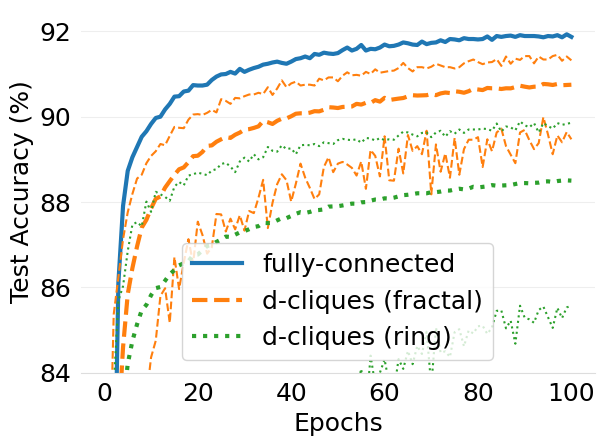}
             \caption{\label{fig:d-cliques-scaling-mnist-1000-linear} Linear}
     \end{subfigure}
     \hfill
     \begin{subfigure}[b]{0.23\textwidth}
         \centering
         \includegraphics[width=\textwidth]{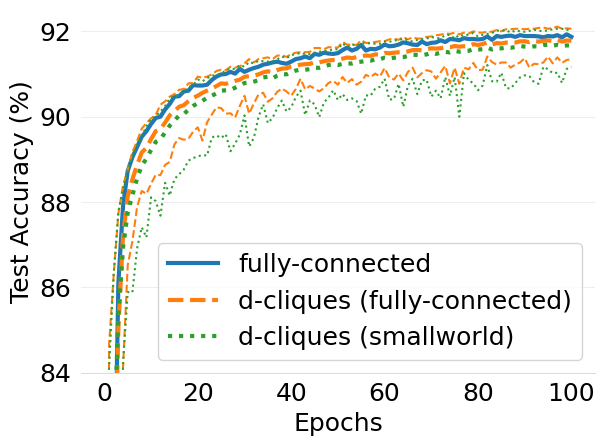}
\caption{\label{fig:d-cliques-scaling-mnist-1000-super-linear}  Super- and Quasi-Linear}
     \end{subfigure}
\caption{\label{fig:d-cliques-scaling-mnist-1000} MNIST: D-Cliques convergence
speed with 1000 nodes (10 nodes per clique, same number of updates per epoch as 100 nodes, i.e. batch-size 10x less per node) and different inter-clique topologies.}
\end{figure}

\begin{figure}[htbp]
     \centering
     \begin{subfigure}[b]{0.23\textwidth}
         \centering
            \includegraphics[width=\textwidth]{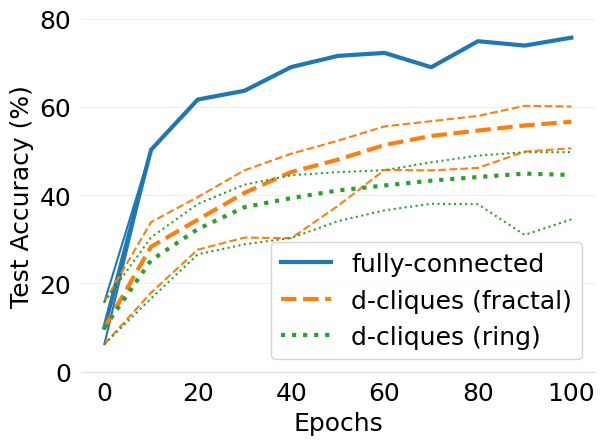}
             \caption{\label{fig:d-cliques-scaling-cifar10-1000-linear} Linear}
     \end{subfigure}
     \hfill
     \begin{subfigure}[b]{0.23\textwidth}
         \centering
         \includegraphics[width=\textwidth]{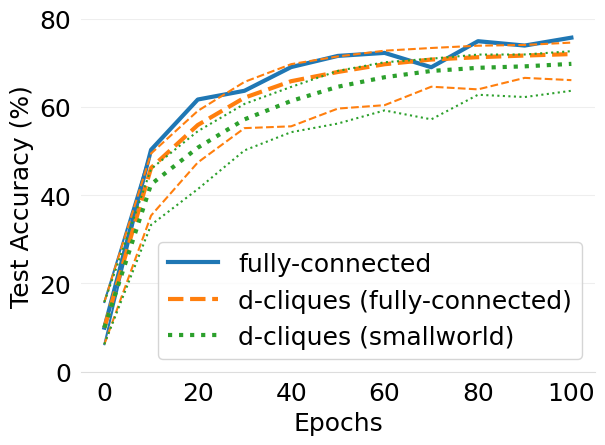}
\caption{\label{fig:d-cliques-scaling-cifar10-1000-super-linear}  Super- and Quasi-Linear}
     \end{subfigure}
\caption{\label{fig:d-cliques-scaling-cifar10-1000} CIFAR10: D-Cliques
convergence speed with 1000 nodes (10 nodes per clique, same number of updates per epoch as 100 nodes, i.e. batch-size 10x less per node) and different inter-clique topologies.}
\end{figure}

In this experiment, we explore the trade-offs between scalability and
convergence speed induced by the several sparse inter-clique topologies
introduced in Section~\ref{section:interclique-topologies}.
\autoref{fig:d-cliques-scaling-mnist-1000} and \autoref{fig:d-cliques-scaling-cifar10-1000}  
show the convergence speed  respectively on MNIST and CIFAR10 on a larger network of 1000 nodes, 
compared to the ideal baseline of a
fully-connected network representing
the fastest convergence speed achievable if topology had no impact. Among the linear schemes, the ring
topology converges but is much slower than our fractal scheme. Among the super-linear schemes, the small-world
topology has a convergence speed that is almost the same as with a
fully-connected inter-clique topology but with 22\% less edges
(14.5 edges on average instead of 18.9).


While the small-world inter-clique topology shows promising scaling behavior, the
fully-connected inter-clique topology still offers
significant benefits with 1000 nodes, as it represents a 98\% reduction in the
number of edges compared to fully connecting individual nodes (18.9 edges on
average instead of 999) and a 96\% reduction in the number of messages (37.8
messages per round per node on average instead of 999). 
We refer to Appendix~\ref{app:scaling} for additional results comparing the convergence speed across different number of nodes. 
Overall, these results show that D-Cliques can gracefully scale with the
number of nodes.

\subsection{Full Intra-Clique Connectivity is Necessary}

\begin{figure}[htbp]
     \centering

\begin{subfigure}[htbp]{0.23\textwidth}
     \centering   
         \includegraphics[width=\textwidth]{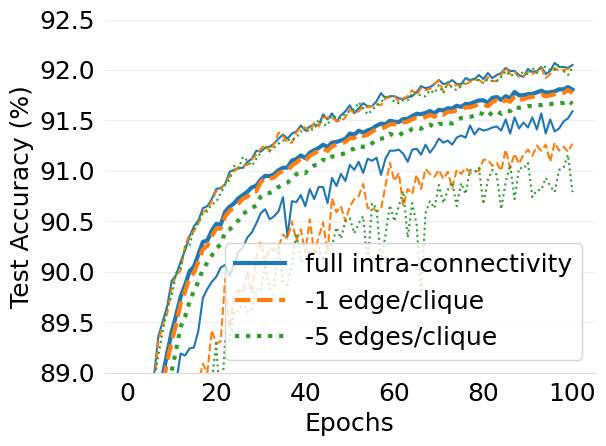}     
\caption{\label{fig:d-cliques-mnist-wo-clique-avg-impact-of-edge-removal} Without Clique Averaging }
\end{subfigure}
\hfill
\begin{subfigure}[htbp]{0.23\textwidth}
     \centering
         \includegraphics[width=\textwidth]{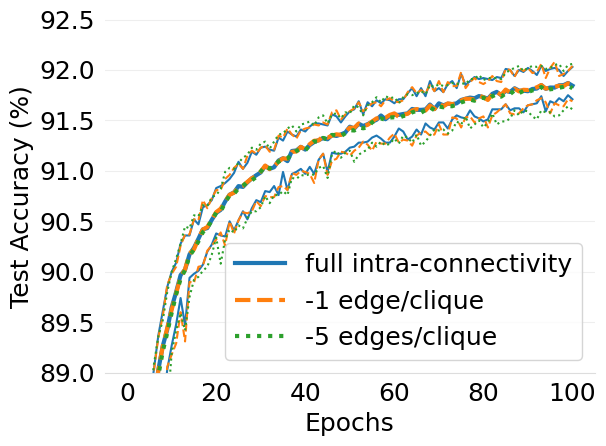}
\caption{\label{fig:d-cliques-mnist-w-clique-avg-impact-of-edge-removal} With Clique Averaging}
\end{subfigure}
\caption{\label{fig:d-cliques-mnist-intra-connectivity} MNIST: Impact of
intra-clique edge removal on D-Cliques (fully-connected) with 10
cliques of 10 heterogeneous nodes (100 nodes). Y axis starts at 89.}
\end{figure}

In this experiment, we measure the impact of removing intra-clique edges 
 to assess how critical full connectivity is within cliques. We choose edges to remove
 among the 45 undirected edges present in cliques of size 10. The removal of
 an edge removes the connection in both directions. We remove 1 and 5 edges
 randomly, respectively 2.2\% and 11\% of intra-clique edges. \autoref{fig:d-cliques-mnist-intra-connectivity} 
 shows that for  MNIST, when not using Clique Averaging, 
removing edges decreases slightly the convergence speed and increases 
the variance between nodes. When using Clique Averaging, removing up to 5
edges does not noticeably affect
the convergence speed and variance.

\begin{figure}[htbp]
     \centering
\begin{subfigure}[htbp]{0.23\textwidth}
     \centering   
         \includegraphics[width=\textwidth]{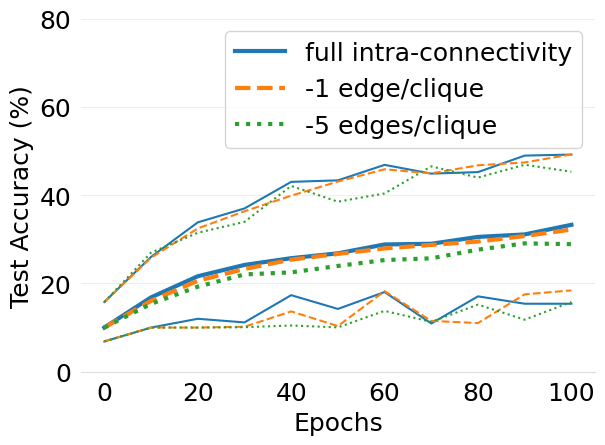}     
\caption{\label{fig:d-cliques-cifar10-wo-clique-avg-impact-of-edge-removal} Without Clique Averaging }
\end{subfigure}
\hfill
\begin{subfigure}[htbp]{0.23\textwidth}
     \centering
         \includegraphics[width=\textwidth]{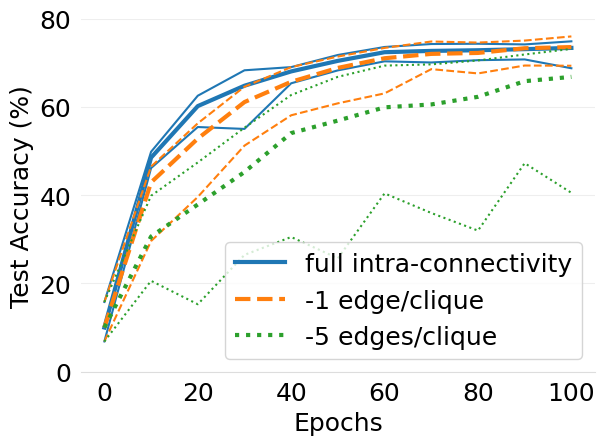}
\caption{\label{fig:d-cliques-cifar10-w-clique-avg-impact-of-edge-removal} With Clique Averaging}
\end{subfigure}
\caption{\label{fig:d-cliques-cifar10-intra-connectivity} CIFAR10: Impact of intra-clique edge removal (with momentum) on D-Cliques (fully-connected) with 10 cliques of 10 heterogeneous nodes (100 nodes).}
\end{figure}

In contrast, \autoref{fig:d-cliques-cifar10-intra-connectivity} shows that for CIFAR10, the impact is stronger. We show the results with and without Clique Averaging
with momentum in both cases, as momentum is critical for obtaining the best
convergence speed on CIFAR10. Without Clique Averaging,
removing edges has a small effect on convergence speed and variance, but the convergence speed is too slow to be practical.
With Clique Averaging, removing a single edge has a small but noticeable
effect. Strikingly, removing 5 edges per clique significantly damages the
convergence and yields a sharp increase in the variance across nodes.
Therefore, while D-Cliques can tolerate the removal of some intra-clique edges
when training simple linear models and datasets as in MNIST, fast
convergence speed and low variance requires full or nearly full connectivity
when using high-capacity models and more difficult datasets. This is
in line with the observations made in Section~\ref{sec:exp:clique_avg}
regarding the effect of Clique Averaging. Again, these results show the
relevance of our design choices, including the choice of constructing fully
connected cliques.

\subsection{Greedy Swap Improves Random Cliques at an Affordable Cost}
\label{section:greedy-swap-vs-random-cliques}

In the next two sub-sections, we compare cliques built with Greedy Swap (Alg.~\ref{Algorithm:greedy-swap})
to Random Cliques, a simple and obvious baseline, on their quality (skew),  the cost 
of their construction, and their convergence speed.

\subsubsection{Cliques with Low Skew can be Constructed Efficiently with Greedy Swap}
\label{section:cost-cliques}

\begin{figure}[htbp]
    \centering
    \includegraphics[width=0.25\textwidth]{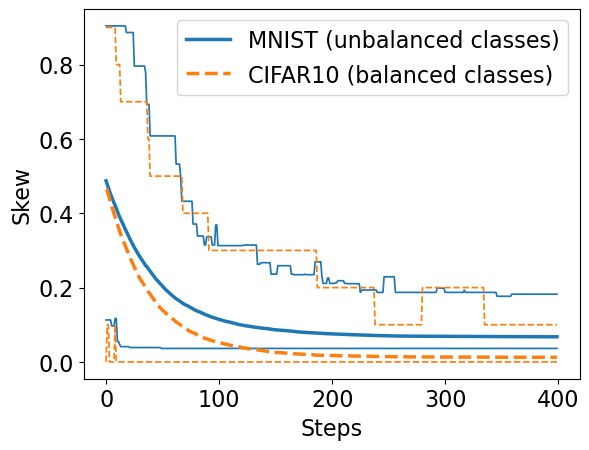}
    \caption{\label{fig:skew-convergence-speed-2-shards} Skew decrease during clique construction of 10 cliques of 10 heterogeneous nodes (100 nodes). Bold line is the average over 100 experiments. Thin lines are respectively the minimum and maximum over all experiments. In wall-clock time, 1000 steps take less than 6 seconds in Python 3.7 on a MacBook Pro 2020.}
\end{figure}

We compared the final average skew of 10 cliques with 10 nodes each (for
$n=100$) created either randomly or with Greedy Swap,
over 100 experiments after 1000 steps. \autoref{fig:skew-convergence-speed-2-shards}, in the form of an histogram,
 shows that Greedy Swap generates cliques of significantly lower skew, close to 0 in a majority of cases for both MNIST and CIFAR10.


\begin{figure}[htbp]
    \centering        
    \begin{subfigure}[b]{0.2\textwidth}
    \centering
    \includegraphics[width=\textwidth]{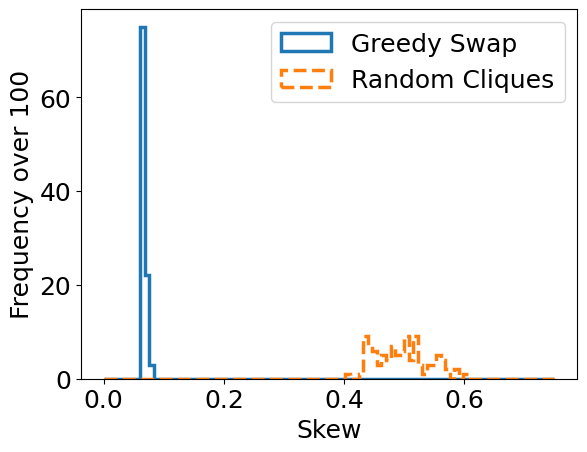}
    \caption{\label{fig:final-skew-distribution-mnist} MNIST }
    \end{subfigure}
    \hfill
    \begin{subfigure}[b]{0.2\textwidth}
    \centering
    \includegraphics[width=\textwidth]{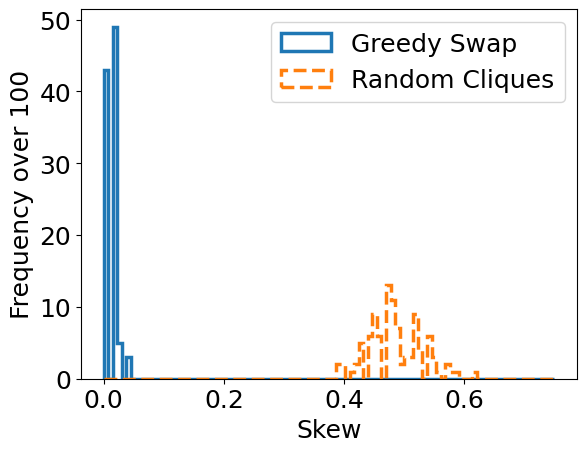}
    \caption{\label{fig:final-skew-distribution-cifar10} CIFAR10}
    \end{subfigure}
\caption{\label{fig:final-skew-distribution} Final quality of cliques (skew) with a maximum size of 10 over 100 experiments in a network of 100 nodes.}
\end{figure}

\autoref{fig:skew-convergence-speed-2-shards} shows such a low skew can be achieved 
in less than 400 steps for both MNIST and CIFAR10. In practice it takes less
than 6 seconds in Python 3.7 on a 
Macbook Pro 2020 for a network of 100 nodes and cliques of size 10. Greedy Swap 
is therefore fast and efficient. Moreover, it illustrates the fact that a
global imbalance in the number of examples
across classes makes the construction of cliques with low skew harder and
slower.

\subsubsection{Cliques built with Greedy Swap Converge Faster than Random Cliques}

\begin{figure}[htbp]
    \centering        
    \begin{subfigure}[b]{0.23\textwidth}
    \centering
    \includegraphics[width=\textwidth]{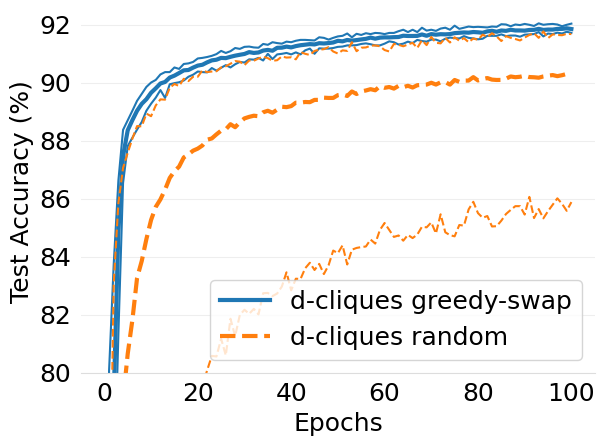}
    \caption{\label{fig:convergence-speed-mnist-dc-random-vs-dc-gs-2-shards-per-node} MNIST}
    \end{subfigure}
    \hfill
    \begin{subfigure}[b]{0.23\textwidth}
    \centering
    \includegraphics[width=\textwidth]{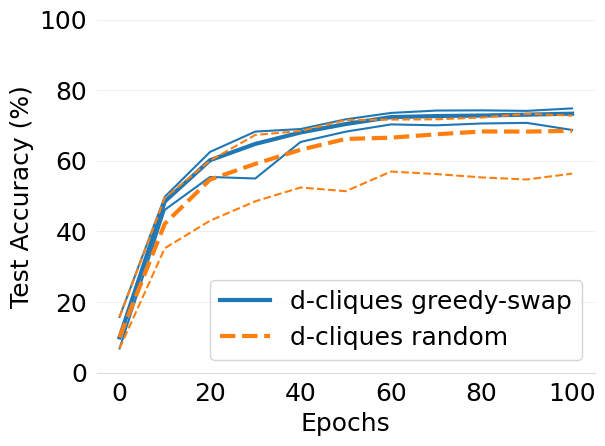}
    \caption{\label{fig:convergence-speed-cifar10-dc-random-vs-dc-gs-2-shards-per-node} CIFAR10}
    \end{subfigure}
\caption{\label{fig:convergence-speed-dc-random-vs-dc-gs-2-shards-per-node} Convergence speed of D-Cliques constructed randomly vs Greedy Swap with 10 cliques of 10 heterogeneous nodes (100 nodes).}
\end{figure}

\autoref{fig:convergence-speed-dc-random-vs-dc-gs-2-shards-per-node} compares
the convergence speed of cliques optimized with Greedy Swap for 1000 steps with cliques built randomly 
(equivalent to Greedy Swap with 0 steps). For both MNIST and CIFAR10, convergence speed
increases significantly and variance between nodes decreases dramatically. Decreasing the skew of cliques
is therefore critical to convergence speed.


\subsection{Additional Experiments on Extreme Label Distribution Skew}

In Appendix~\ref{app:extreme-local-skew}, we replicate experimental
results on an extreme case of label distribution skew where each node only has
examples of a single class. These results consistently show that our
approach remains effective even for extremely skewed label distributions
across nodes.


\section{Related Work}
\label{section:related-work}

In this section, we review some related work on dealing with heterogeneous
data in federated learning, and on the role of topology in fully decentralized
algorithms.

\paragraph{Dealing with heterogeneity in server-based FL.}
Data heterogeneity is not much of an issue in server-based FL if
clients send their parameters to the server after each gradient update.
Problems arise when one seeks to reduce
the number of communication rounds by allowing each participant to perform
multiple local updates, as in the popular FedAvg algorithm 
\cite{mcmahan2016communication}. Indeed, data heterogeneity can prevent
such algorithms from
converging to a good solution \cite{quagmire,scaffold}. This led to the design
of algorithms that are specifically designed to mitigate the impact
of heterogeneity while performing
multiple local updates, using adaptive client sampling \cite{quagmire}, update
corrections \cite{scaffold} or regularization in the local objective 
\cite{fedprox}. Another direction is to embrace the heterogeneity by
learning personalized models for each client 
\cite{smith2017federated,perso_fl_mean,maml,moreau,Marfoq2021a}.
We note that recent work explores rings of server-based topologies 
\cite{tornado}, but the focus is not on dealing with heterogeneous data but
to make server-based FL more scalable to a large number of clients.

\paragraph{Dealing with heterogeneity in fully decentralized FL.}
Data heterogeneity is known to negatively impact the convergence speed
of fully decentralized FL algorithms in practice \cite{jelasity}. Aside from approaches that aim to learn personalized models \cite{Vanhaesebrouck2017a,Zantedeschi2020a}, this
motivated the design of algorithms with modified updates based on variance
reduction \cite{tang18a}, momentum correction \cite{momentum_noniid},
cross-gradient
aggregation \cite{cross_gradient}, or multiple averaging steps
between updates \citep[see][and references therein]{consensus_distance}. These
algorithms
typically require significantly more communication and/or computation, and
have only been evaluated on small-scale networks with a few tens of
nodes.\footnote{We
also observed that \cite{tang18a} is subject to numerical
instabilities when run on topologies other than rings. When
the rows and columns of $W$ do not exactly
sum to $1$ (due to finite precision), these small differences get amplified by
the proposed updates and make the algorithm diverge.}
In contrast, D-Cliques focuses on the design of a sparse topology which is
able to compensate for the effect of heterogeneous data and scales to large
networks. We do not modify the simple
and efficient D-SGD
algorithm \cite{lian2017d-psgd} beyond removing some neighbor
contributions
that otherwise bias the gradient direction.

\paragraph{Impact of topology in fully decentralized FL.} It is well
known
that the choice of network topology can affect the
convergence of fully decentralized algorithms. In theoretical convergence
rates, this is typically accounted
for by a dependence on the spectral gap of
the network, see for instance 
\cite{Duchi2012a,Colin2016a,lian2017d-psgd,Nedic18}.
However, for homogeneous (IID) data, practice contradicts these classic
results as fully decentralized algorithms have been observed to converge
essentially as fast
on sparse topologies like rings or grids as they do on a fully connected
network \cite{lian2017d-psgd,Lian2018}. Recent work 
\cite{neglia2020,consensus_distance} sheds light on this phenomenon with refined convergence analyses based on differences between gradients or parameters across nodes, which are typically
smaller in the homogeneous case. However, these results do not give any clear insight
regarding the role of the topology in the presence of heterogeneous data. 
We note that some work
has gone into designing efficient topologies to optimize the use of
network resources \citep[see e.g.,][]{marfoq}, but the topology is chosen
independently of how data is distributed across nodes. In summary, the role
of topology in the heterogeneous data scenario is not well understood and we are not
aware of prior work focusing on this question. Our work is the first
to show that an
appropriate choice of data-dependent topology can effectively compensate for
heterogeneous data.

\section{Conclusion}
\label{section:conclusion}

We proposed D-Cliques, a sparse topology that obtains similar convergence
speed as a fully-connected network in the presence of label distribution skew.
D-Cliques is based on assembling subsets of nodes into cliques such
that the clique-level class distribution is representative of the global
distribution, thereby locally recovering homogeneity of data. Cliques are
connected together by a
sparse inter-clique topology so that
they quickly converge to the same model. We proposed Clique
Averaging to remove the bias in gradient computation due to non-homogeneous
averaging neighborhood by averaging gradients only with other nodes within the clique. Clique Averaging
can in turn be used to implement an effective momentum.
Through our extensive set of experiments, we
showed that the clique structure of D-Cliques is critical in obtaining these
results and that a small-world inter-clique topology with only $O(n \log n)$ 
edges achieves a very good compromise between
convergence speed and scalability with the number of nodes.

D-Cliques thus appears to be very promising to reduce bandwidth
usage on FL servers and to implement fully decentralized alternatives in a
wider range of applications where global coordination is impossible or costly.
For instance, the relative frequency of classes in each node
could be computed using PushSum~\cite{kempe2003gossip}, and the topology could
be constructed in a decentralized and adaptive way with
PeerSampling~\cite{jelasity2007gossip}. This will be investigated in future work.
We also believe that our ideas can be useful to deal
with more general types of data heterogeneity beyond the important case
of
label distribution skew on which we focused in this paper. An important
example is
covariate shift or feature distribution skew \cite{kairouz2019advances}, for
which local density estimates could be used as basis to construct cliques that
approximately recover the global distribution.

\bibliography{main.bib}
\bibliographystyle{mlsys2022}


\appendix

\newpage
~
\newpage

\section{Details on Small-world Inter-clique Topology}
\label{app:small_world}

 We present a more detailed and precise explanation of the algorithm to establish a small-world
 inter-clique topology (Algorithm~\ref{Algorithm:Smallworld}). Algorithm~\ref{Algorithm:Smallworld} instantiates the function 
\texttt{inter} with a
small-world inter-clique topology as described in Section~\ref{section:interclique-topologies}. It adds a
linear number of inter-clique edges by first arranging cliques on a ring. It then adds a logarithmic number of ``finger'' edges to other cliques on the ring chosen such that there is a constant number of edges added per set, on sets that are exponentially bigger the further away on the ring. ``Finger'' edges are added symmetrically on both sides of the ring to the cliques in each set that are closest to a given set. ``Finger`` edges are added for each clique on the ring, therefore adding in total a linear-logarithmic number of edges.

\begin{algorithm}[h]
   \caption{$\textit{smallworld}(DC)$:  adds $O(\# N \log(\# N))$ edges}
   \label{Algorithm:Smallworld}
   \begin{algorithmic}[1]
        \STATE \textbf{Require:} set of cliques $DC$ (set of set of nodes)
        \STATE ~~size of neighborhood $ns$ (default 2)
        \STATE ~~function $\textit{least\_edges}(S, E)$ that returns one of the nodes in $S$ with the least number of edges in $E$
        \STATE $E \leftarrow \emptyset$ \COMMENT{Set of Edges}
        \STATE $L \leftarrow [ C~\text{for}~C \in DC ]$ \COMMENT{Arrange cliques in a list}
        \FOR{$i \in \{1,\dots,\#DC\}$} 
          \FOR{$\textit{offset} \in \{ 2^x~\text{for}~x~\in \{ 0, \dots, \lceil \log_2(\#DC) \rceil \} \}$} 
             \FOR{$k \in \{0,\dots,ns-1\}$}
                 \STATE $n \leftarrow \textit{least\_edges}(L_i, E)$
                 \STATE $m \leftarrow \textit{least\_edges}(L_{(i+\textit{offset}+k) \% \#DC}, E)$ 
                 \STATE $E \leftarrow E \cup \{ \{n,m\} \}$
                 \STATE $n \leftarrow \textit{least\_edges}(L_i, E)$
                 \STATE $m \leftarrow \textit{least\_edges}(L_{(i-\textit{offset}-k)\% \#DC} , E)$ 
                 \STATE $E \leftarrow E \cup \{ \{n,m\} \}$
             \ENDFOR
           \ENDFOR
        \ENDFOR
         \RETURN E
   \end{algorithmic}
\end{algorithm}

Algorithm~\ref{Algorithm:Smallworld}  expects a set of cliques $DC$, previously computed by 
Algorithm~\ref{Algorithm:greedy-swap}; a size of neighborhood $ns$,
which is the number of finger edges to add per set of cliques, and a function 
\textit{least\_edges}, which given a set of nodes $S$ and an existing set of
edges $E =  \{\{i,j\}, \dots \}$, returns one of the nodes in $E$ with the least number of edges. It returns a new set of edges $\{\{i,j\}, \dots \}$ with all edges added by the small-world topology.

The implementation first arranges the cliques of $DC$ in a list, which
represents the ring. Traversing the list with increasing indices is equivalent
to traversing the ring in the clockwise direction, and inversely. Then, for every clique $i$ on the ring from which we are computing the distance to others, a number of edges are added. All other cliques are implicitly arranged in mutually exclusive sets, with size and at offset exponentially bigger (doubling at every step). Then for every of these sets, $ns$ edges are added, both in the clockwise and counter-clockwise directions, always on the nodes with the least number of edges in each clique. The ring edges are implicitly added to the cliques at offset $1$ in both directions.

\section{Additional Experiments on Scaling Behavior with Increasing Number of
Nodes}
\label{app:scaling}

Section~\ref{section:scaling} compares the convergence speed of various inter-clique topologies at a scale of 1000 nodes. In this section, we show the effect of scaling the number of nodes, by comparing the convergence speed with 1, 10, 100, and 1000 nodes, and adjusting the batch size to maintain a constant number of updates per epoch. We present results for Ring, Fractal, Small-world, and Fully-Connected inter-clique topologies.
 
\begin{figure}[htbp]
         \centering     
      \begin{subfigure}[b]{0.35\textwidth}
         \centering
         \includegraphics[width=\textwidth]{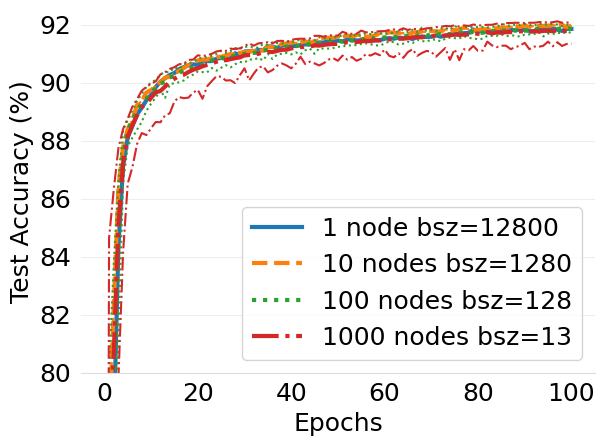}
         \caption{Fully-Connected}
     \end{subfigure}
     \quad
      \begin{subfigure}[b]{0.35\textwidth}
         \centering
         \includegraphics[width=\textwidth]{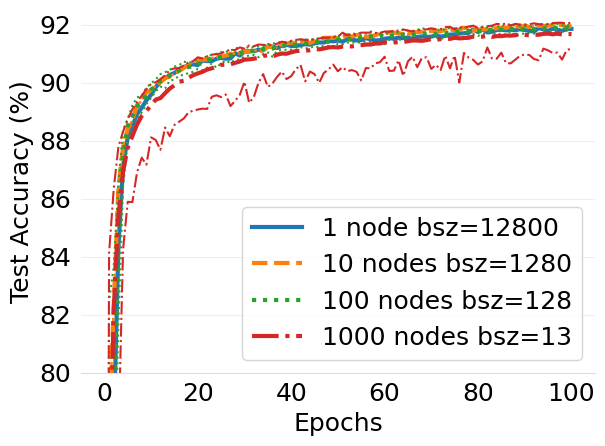}
         \caption{Small-world}
     \end{subfigure}
     \quad

         \begin{subfigure}[b]{0.35\textwidth}
         \centering
         \includegraphics[width=\textwidth]{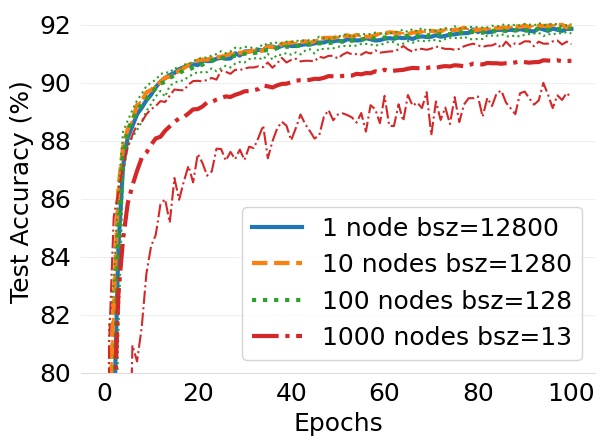}
         \caption{Fractal}
     \end{subfigure}  
     \quad
         \begin{subfigure}[b]{0.35\textwidth}
         \centering
         \includegraphics[width=\textwidth]{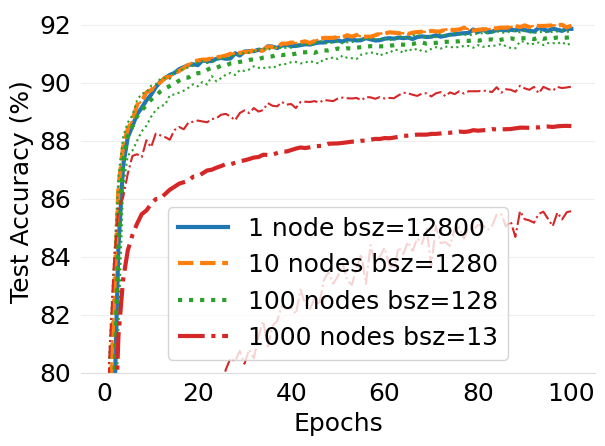}
         \caption{Ring}
     \end{subfigure}  
     
     \caption{\label{fig:d-cliques-mnist-scaling-fully-connected} MNIST:
     D-Cliques scaling behavior (constant updates per epoch and 10 nodes per clique) for different
     inter-clique topologies.} 
\end{figure}

Figure~\ref{fig:d-cliques-mnist-scaling-fully-connected} shows the results for
MNIST. For all topologies, we notice a perfect scaling up to 100 nodes, i.e.
the accuracy curves overlap, with low variance between nodes. Starting at 1000
nodes, there is a significant increase in variance between nodes and the
convergence is slower, only marginally for Fully-Connected but
significantly so for Fractal and Ring. Small-world has higher variance between nodes but maintains a convergence speed close to that of Fully-Connected.


\begin{figure}[htbp]
         \centering
      \begin{subfigure}[b]{0.35\textwidth}
         \centering
         \includegraphics[width=\textwidth]{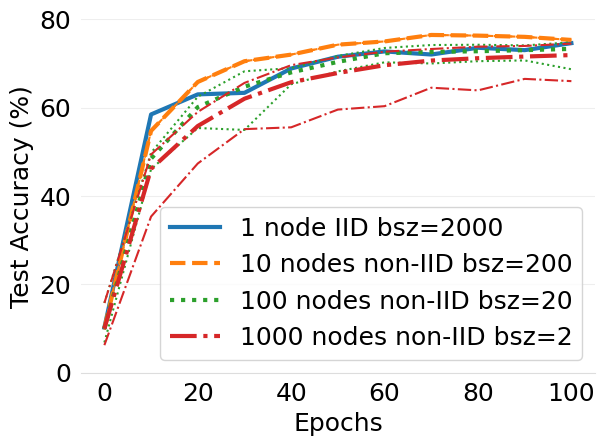}
         \caption{Fully-Connected}
     \end{subfigure}
     \quad
      \begin{subfigure}[b]{0.35\textwidth}
         \centering
         \includegraphics[width=\textwidth]{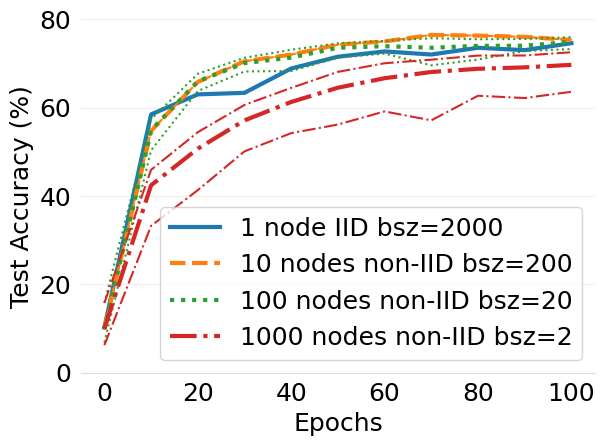}
         \caption{Small-world}
     \end{subfigure}

         \begin{subfigure}[b]{0.35\textwidth}
         \centering
         \includegraphics[width=\textwidth]{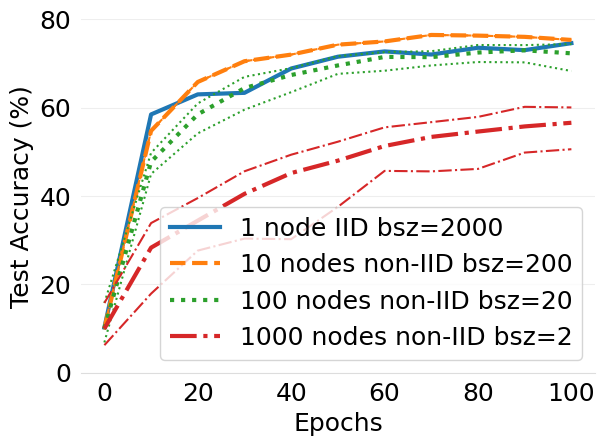}
         \caption{Fractal}
     \end{subfigure}  
     \quad

         \begin{subfigure}[b]{0.35\textwidth}
         \centering
         \includegraphics[width=\textwidth]{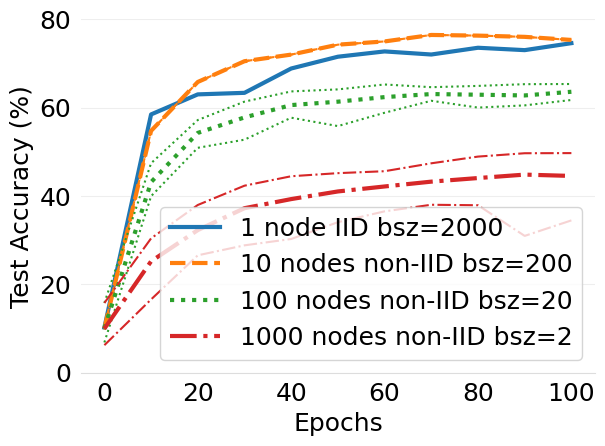}
         \caption{Ring}
     \end{subfigure}  
     
     \caption{\label{fig:d-cliques-cifar10-scaling-fully-connected} CIFAR10: D-Cliques scaling behavior (constant updates per epoch and 10 nodes per clique) for different
     inter-clique topologies.} 
\end{figure}

Figure~\ref{fig:d-cliques-cifar10-scaling-fully-connected} shows the results
for CIFAR10. When increasing from 1 to 10 nodes (resulting in a single
fully-connected clique), there is actually a small increase both in final
accuracy and convergence speed. We believe this increase is due to the
gradient being computed with better representation of examples from all
classes with 10 fully-connected non-IID nodes, while the gradient for a single
non-IID node may have a slightly larger bias because the random sampling 
may allow more bias in the representation of classes in each batch. At a
scale of 100 nodes, there is no difference between Fully-Connected and
Fractal, as the connections are the same; however, a Ring already shows a
significantly slower convergence. At 1000 nodes, the convergence significantly
slows down for Fractal and Ring, while remaining close, albeit with a larger
variance, to Fully-Connected. Similar to MNIST, Small-world has
higher variance and slightly lower convergence speed than Fully-Connected but
remains very close.

We therefore conclude that Fully-Connected and Small-world have good scaling
properties in terms of convergence speed, and that the
linear-logarithmic number of edges of Small-world makes it the best compromise
between convergence speed and connectivity, and thus the best choice for
efficient large-scale decentralized learning in practice.


\section{Additional Experiments with Extreme Label Skew}
\label{app:extreme-local-skew} 

In this section, we present additional results for similar experiments as in
Section~\ref{section:evaluation} but in the presence of
 \textit{extreme label distribution skew}: we consider that each node only has examples from a single class. This extreme partitioning case provides an upper bound on the effect of label distribution skew suggesting that D-Cliques should perform similarly or better in less extreme cases, as long as a small-enough average skew can be obtained on all cliques. In turn, this helps to provide insights on why D-Cliques work well, as well as to quantify the loss in convergence speed
that may result from using construction algorithms that generate cliques with higher skew.

\subsection{Data Heterogeneity Assumptions}
\label{section:non-iid-assumptions}

To isolate the effect of label distribution skew from other potentially compounding
factors, we make the following simplifying assumptions: (1) All classes are
equally represented in the global dataset; (2) All classes are represented on
the same number of nodes; (3) All nodes have the same number of examples.

While less realistic than the assumptions used Section~\ref{section:evaluation}, 
these assumptions are still reasonable because: (1) Global class imbalance equally
affects the optimization process on a single node and is therefore not
specific to the decentralized setting; (2) Our results do not exploit specific
positions in the topology;  (3) Imbalanced dataset sizes across nodes can be
addressed for instance by appropriately weighting the individual loss
functions.

These assumptions do make the construction of cliques slightly easier by 
making it easy to build cliques that have zero skew, as shown in 
Section~\ref{section:ideal-cliques}. 

\subsection{Constructing Ideal Cliques}
\label{section:ideal-cliques}
 
 Algorithm~\ref{Algorithm:D-Clique-Construction} shows the overall approach
 for constructing a D-Cliques topology under the assumptions of Section~\ref{section:non-iid-assumptions}.\footnote{An IID
 version of D-Cliques, in which each node has an equal number of examples of
 all classes, can be implemented by picking $\#L$ nodes per clique at random.}
 It expects the following inputs: $L$, the set of all classes present in the global distribution $D = \bigcup_{i \in N} D_i$; $N$, the set of all nodes; a function $classes(S)$, which given a subset $S$ of nodes in $N$ returns the set of classes in their joint local distributions ($D_S = \bigcup_{i \in S} D_i$); a function $intra(DC)$, which given $DC$, a set of cliques (set of set of nodes), creates a set of edges ($\{\{i,j\}, \dots \}$) connecting all nodes within each clique to one another; a function $inter(DC)$, which given a set of cliques, creates a set of edges ($\{\{i,j\}, \dots \}$) connecting nodes belonging to different cliques; and a function $weigths(E)$, which given a set of edges, returns the weighted matrix $W_{ij}$.  Algorithm~\ref{Algorithm:D-Clique-Construction} returns both $W_{ij}$, for use in D-SGD (Algorithm~\ref{Algorithm:D-PSGD} and~\ref{Algorithm:Clique-Unbiased-D-PSGD}), and $DC$, for use with Clique Averaging (Algorithm~\ref{Algorithm:Clique-Unbiased-D-PSGD}).
 
   \begin{algorithm}[h]
   \caption{D-Cliques Construction}
   \label{Algorithm:D-Clique-Construction}
   \begin{algorithmic}[1]
        \STATE \textbf{Require:} set of classes globally present $L$, 
        \STATE~~ set of all nodes $N = \{ 1, 2, \dots, n \}$,
        \STATE~~ fn $\textit{classes}(S)$ that returns the classes present in a subset of nodes $S$,
        \STATE~~ fn $\textit{intra}(DC)$ that returns edges intraconnecting cliques of $DC$,
        \STATE~~ fn $\textit{inter}(DC)$ that returns edges interconnecting cliques of $DC$ (Sec.~\ref{section:interclique-topologies})
         \STATE~~ fn $\textit{weights}(E)$ that assigns weights to edges in $E$ 
         
        \STATE $R \leftarrow \{ n~\text{for}~n \in N \}$ \COMMENT{Remaining nodes}
        \STATE $DC \leftarrow \emptyset$ \COMMENT{D-Cliques}
        \STATE $\textit{C} \leftarrow \emptyset$ \COMMENT{Current Clique}
        \WHILE{$R \neq \emptyset$}
       \STATE $n \leftarrow \text{pick}~1~\text{from}~\{ m \in R | \textit{classes}(\{m\}) \subsetneq \textit{classes}(\textit{C}) \}$
       \STATE $R \leftarrow R \setminus \{ n \}$
       \STATE $C \leftarrow C \cup \{ n \}$
       \IF{$\textit{classes}(C) = L$}
           \STATE $DC \leftarrow DC \cup \{ C \}$
           \STATE $C \leftarrow \emptyset$
       \ENDIF
        \ENDWHILE
        \RETURN $(weights(\textit{intra}(DC) \cup \textit{inter}(DC)), DC)$
   \end{algorithmic}
\end{algorithm}
 
The implementation builds a single clique by adding nodes with different
classes until all classes of the global distribution are represented. Each
clique is built sequentially until all nodes are parts of cliques.
Because all classes are represented on an equal number of nodes, all cliques
will have nodes of all classes. Furthermore, since nodes have examples
of a single class, we are guaranteed a valid assignment is possible in a greedy manner. 
After cliques are created, edges are added and weights are assigned to edges, 
using the corresponding input functions.

\subsection{Evaluation}
\label{section:ideal-cliques-evaluation}

In this section, we provide figures analogous to those of the main text using the partitioning 
scheme of Section~\ref{section:non-iid-assumptions}.

\subsubsection{Data Heterogeneity is Significant at Multiple Levels of Node Skew} 

\begin{figure*}[htbp]
     \centering
     \begin{subfigure}[b]{0.25\textwidth}
         \centering
         \includegraphics[width=\textwidth]{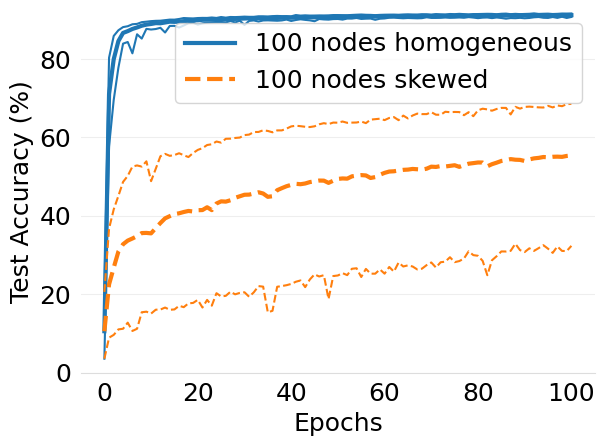}
\caption{\label{fig:ring-IID-vs-non-IID-eq-classes-1-class-per-node} Ring topology}
     \end{subfigure}
     \quad
     \begin{subfigure}[b]{0.25\textwidth}
         \centering
         \includegraphics[width=\textwidth]{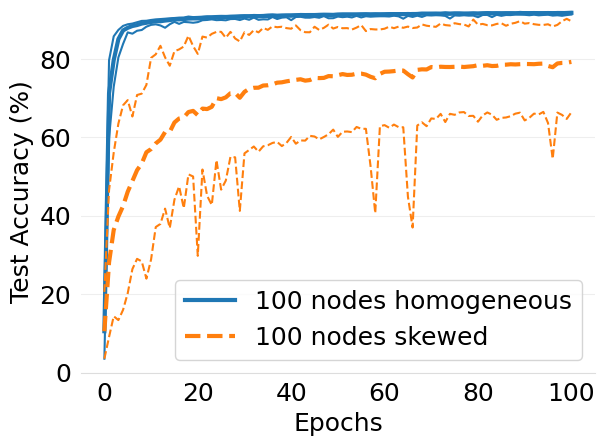}
\caption{\label{fig:grid-IID-vs-non-IID-eq-classes-1-class-per-node} Grid topology}
     \end{subfigure}
     \quad
     \begin{subfigure}[b]{0.25\textwidth}
         \centering
         \includegraphics[width=\textwidth]{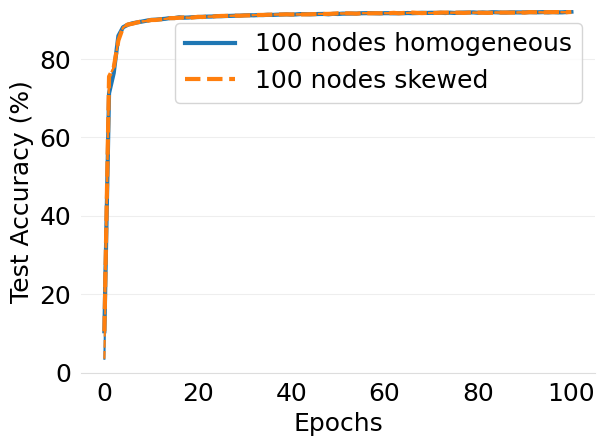}
\caption{\label{fig:fully-connected-IID-vs-non-IID-eq-classes-1-class-per-node} Fully-connected topology}
     \end{subfigure}
        \caption{Convergence speed of decentralized SGD with and without label distribution skew for different topologies on MNIST (Variation of \autoref{fig:iid-vs-non-iid-problem} using balanced classes and skewed with 1 class/node).
        \label{fig:iid-vs-non-iid-problem-1-class-per-node}}
\end{figure*}

\autoref{fig:iid-vs-non-iid-problem-1-class-per-node} is consistent  with \autoref{fig:iid-vs-non-iid-problem} albeit
with slower convergence speed and higher variance.  On the one hand, \autoref{fig:iid-vs-non-iid-problem-1-class-per-node} shows that an extreme  skew amplifies the difficulty of learning. On the other hand, \autoref{fig:iid-vs-non-iid-problem} shows that the problem is not limited to the most extreme cases and is therefore worthy of consideration in designing decentralized federated learning solutions.



\subsubsection{D-Cliques Match the Convergence Speed of Fully-Connected with a Fraction of the Edges}

\begin{figure}[htbp]
    \centering        
    \begin{subfigure}[b]{0.23\textwidth}
    \centering
    \includegraphics[width=\textwidth]{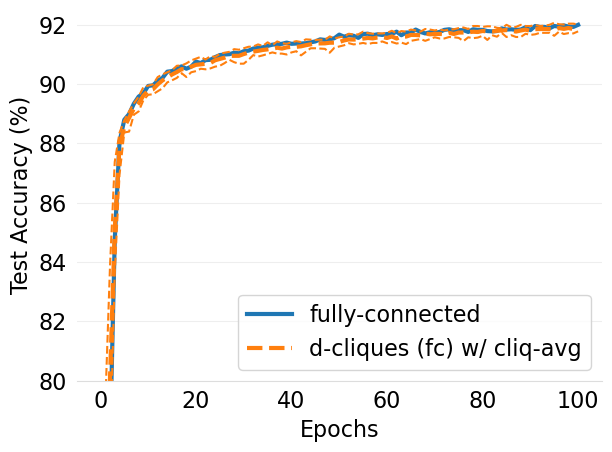}
    \caption{\label{fig:convergence-speed-mnist-dc-fc-vs-fc-1-class-per-node} MNIST}
    \end{subfigure}
    \hfill
    \begin{subfigure}[b]{0.23\textwidth}
    \centering
    \includegraphics[width=\textwidth]{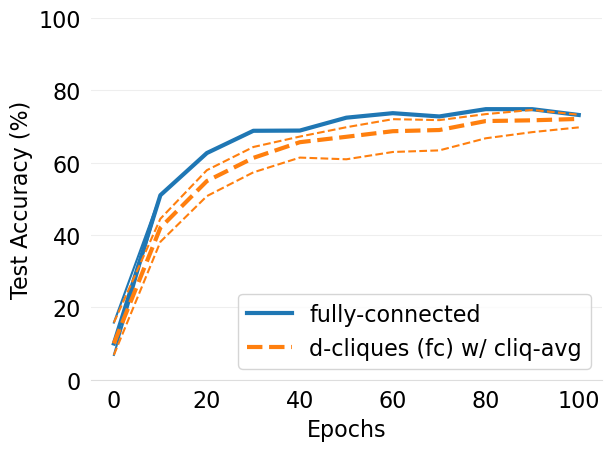}
    \caption{\label{fig:convergence-speed-cifar10-dc-fc-vs-fc-1-class-per-node} CIFAR10 (with momentum)}
    \end{subfigure}
\caption{\label{fig:convergence-speed-dc-vs-fc-1-class-per-node} Comparison on 100 heterogeneous nodes
between a fully-connected network and D-Cliques (fully-connected) constructed with Greedy Swap (10 cliques of 10 nodes) using
Clique Averaging. (Variation of \autoref{fig:convergence-speed-dc-vs-fc-2-shards-per-node} with 1 class/node instead of 2 shards/node).}
\end{figure}

\autoref{fig:convergence-speed-dc-vs-fc-1-class-per-node} shows consistent
results with \autoref{fig:convergence-speed-dc-vs-fc-2-shards-per-node}:
D-Cliques work equally well in more extreme skew. It should therefore work
well for other levels of label distribution skew commonly encountered in
practice.


\subsubsection{Clique Averaging and Momentum are Beneficial and Sometimes Necessary}

\begin{figure}[htbp]
         \centering
         \includegraphics[width=0.23\textwidth]{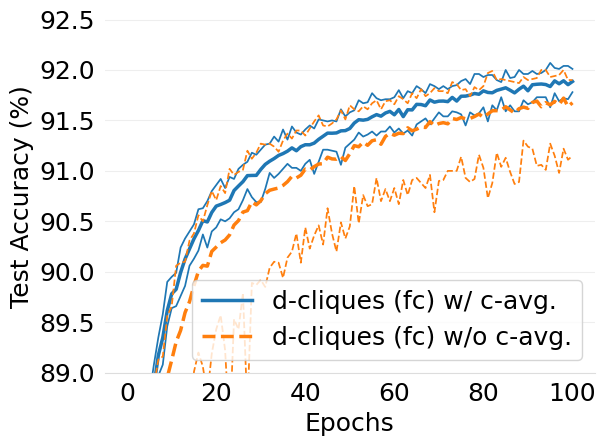}
\caption{\label{fig:d-clique-mnist-clique-avg-1-class-per-node} MNIST: Effect of Clique Averaging on D-Cliques (fully-connected) with 10 cliques of 10 heterogeneous nodes (100 nodes). Y axis starts at 89. (Variation of \autoref{fig:d-clique-mnist-clique-avg} with balanced classes and 1 class/node instead of 2 shards/node).}
\end{figure}

\begin{figure}[htbp]
    \centering        
    \begin{subfigure}[b]{0.23\textwidth}
    \centering
    \includegraphics[width=\textwidth]{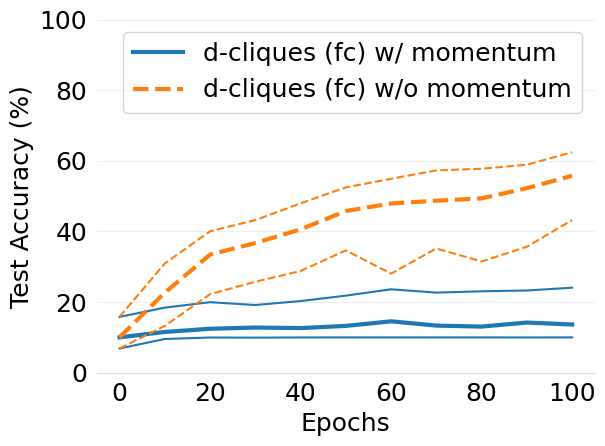}
    \caption{\label{fig:convergence-speed-cifar10-wo-c-avg-no-mom-vs-mom-1-class-per-node} Without Clique Averaging }
    \end{subfigure}
    \hfill
    \begin{subfigure}[b]{0.23\textwidth}
    \centering
    \includegraphics[width=\textwidth]{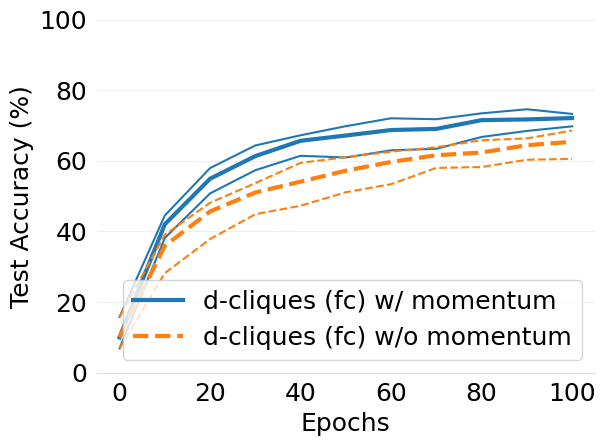}
    \caption{\label{fig:convergence-speed-cifar10-w-c-avg-no-mom-vs-mom-1-class-per-node} With Clique Averaging}
    \end{subfigure}
\caption{\label{fig:cifar10-c-avg-momentum-1-class-per-node} CIFAR10: Effect of Clique Averaging, without and with
momentum, on D-Cliques (fully-connected) with 10 cliques of 10 heterogeneous nodes (100 nodes) (variation of \autoref{fig:cifar10-c-avg-momentum} with 1 class/node instead of 2 shards/node).}
\end{figure}

\autoref{fig:d-clique-mnist-clique-avg-1-class-per-node} and \autoref{fig:cifar10-c-avg-momentum-1-class-per-node} show that, compared respectively to \autoref{fig:d-clique-mnist-clique-avg} and \autoref{fig:cifar10-c-avg-momentum}, Clique Averaging increases in importance the more extreme the skew is and provides consistent convergence speed at multiple levels.

\subsubsection{D-Cliques Clustering is Necessary}
\label{section:d-cliques-clustering-is-necessary}

In this experiment, we compare D-Cliques to different variations of random graphs,
with additional variations compared to the experiments of Section~\ref{section:d-cliques-vs-random-graphs}, 
to show it is actually necessary. Compared to a random graph, D-Cliques enforce additional constraints 
and provide additional mechanisms: they ensure
a diverse representation of all classes in the immediate neighbourhood of all nodes; they enable
 Clique Averaging to debias gradients; and they provide a high-level of clustering, i.e. neighbors 
 of a node are neighbors themselves, which tends to lower variance.
In order to distinguish the effect of the first two from the last, we compare D-Cliques to other variations 
of random graphs: (1) with the additional constraint that all classes should be represented in the immediate neighborhood of all nodes 
(i.e. 'diverse neighbors'), and (2) in combination with unbiased gradients computed using 
the average of the gradients of a subset of neighbors of a node such that the skew of that subset is 0.

The partitioning scheme we use (Section~\ref{section:non-iid-assumptions}) makes the construction of both D-Cliques and diverse random graphs easy and ensures that in both cases the skew of the cliques or neighborhood subset is exactly 0. This removes the challenge of designing topology optimization algorithms for both D-Cliques and random graphs that would guarantee reaching the same level of skews in both cases to make results comparable.

\begin{figure}[htbp]
     \centering     
         \begin{subfigure}[b]{0.23\textwidth}
         \centering
         \includegraphics[width=\textwidth]{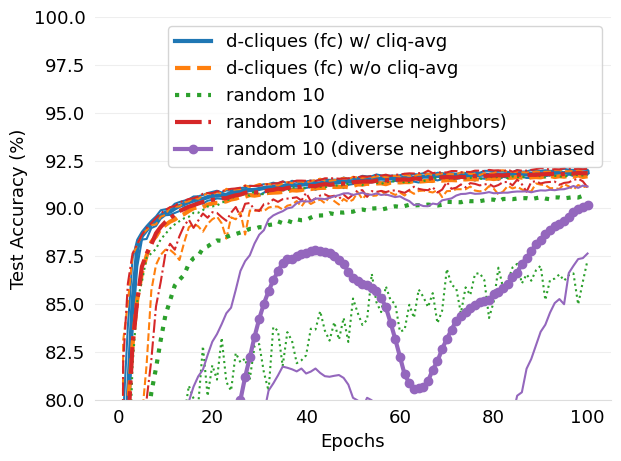}
                  \caption{MNIST}
         \end{subfigure}
                 \hfill                      
        \begin{subfigure}[b]{0.23\textwidth}
        \centering
         \includegraphics[width=\textwidth]{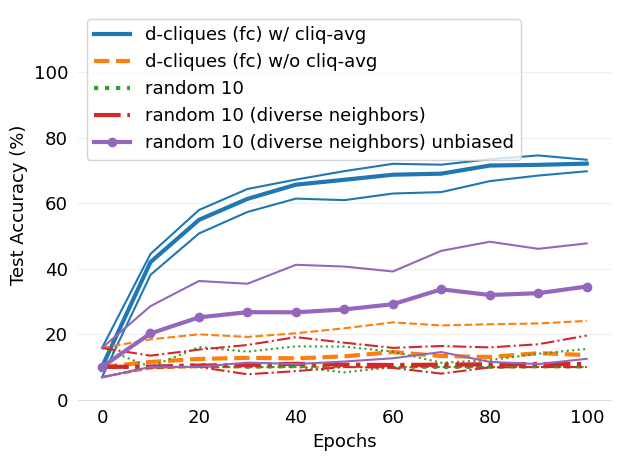}
         \caption{CIFAR10}
     \end{subfigure} 
 \caption{\label{fig:convergence-random-vs-d-cliques-1-class-per-node} Comparison to variations of Random Graph with 10 edges per node on 100 nodes (variation of \autoref{fig:convergence-random-vs-d-cliques-2-shards} with 1 class/node instead of 2 shards/node as well as additional random graphs with more constraints).} 
\end{figure}

\autoref{fig:convergence-random-vs-d-cliques-1-class-per-node} compares the convergence speed of D-Cliques with all the variations of random graphs on both MNIST and CIFAR10. In both cases,
D-Cliques converge faster than all other options. In addition, in the case of CIFAR10, the clustering appears to be critical
for good convergence speed: even a random graph with diverse neighborhoods and unbiased gradients 
converges significantly slower.

\subsubsection{D-Cliques Scale with Sparser Inter-Clique Topologies}

\begin{figure}[htbp]
     \centering
     \begin{subfigure}[b]{0.23\textwidth}
         \centering
            \includegraphics[width=\textwidth]{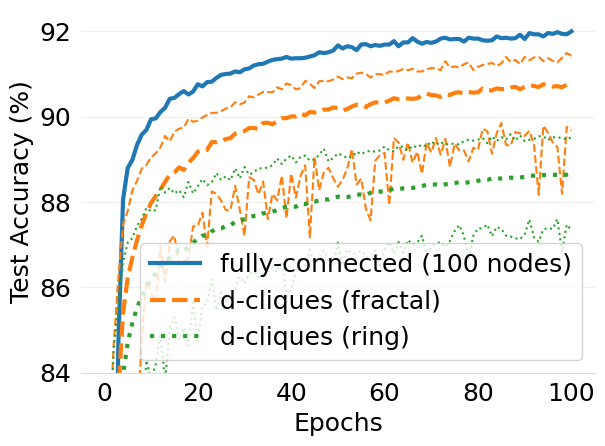}
             \caption{\label{fig:d-cliques-scaling-mnist-1000-linear-1-class-per-node} Linear}
     \end{subfigure}
     \hfill
     \begin{subfigure}[b]{0.23\textwidth}
         \centering
         \includegraphics[width=\textwidth]{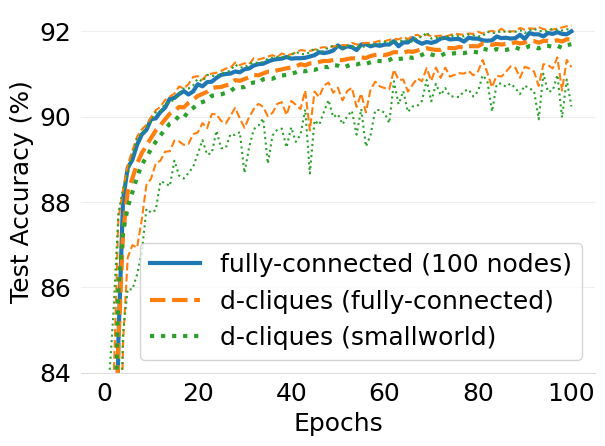}
\caption{\label{fig:d-cliques-scaling-mnist-1000-super-linear-1-class-per-node}  Super- and Quasi-Linear}
     \end{subfigure}
\caption{\label{fig:d-cliques-scaling-mnist-1000-1-class-per-node} MNIST: D-Cliques convergence speed with 1000 nodes (10 nodes per clique, same number of updates per epoch as 100 nodes, i.e. batch-size 10x less per node) with different inter-clique topologies. (variation of \autoref{fig:d-cliques-scaling-mnist-1000} with 1 class/node instead of 2 shards/node).}
\end{figure}

\begin{figure}[htbp]
     \centering
     \begin{subfigure}[b]{0.23\textwidth}
         \centering
            \includegraphics[width=\textwidth]{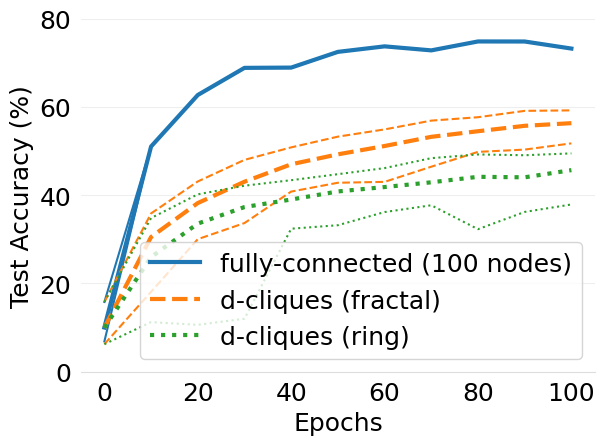}
             \caption{\label{fig:d-cliques-scaling-cifar10-1000-linear-1-class-per-node} Linear}
     \end{subfigure}
     \hfill
     \begin{subfigure}[b]{0.23\textwidth}
         \centering
         \includegraphics[width=\textwidth]{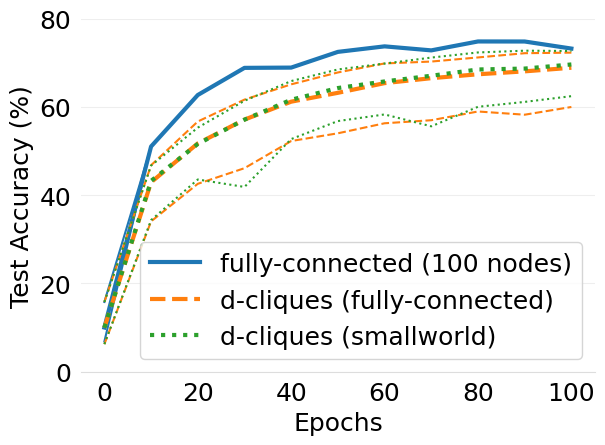}
\caption{\label{fig:d-cliques-scaling-cifar10-1000-super-linear-1-class-per-node}  Super- and Quasi-Linear}
     \end{subfigure}
\caption{\label{fig:d-cliques-scaling-cifar10-1000-1-class-per-node} CIFAR10: D-Cliques convergence speed with 1000 nodes (10 nodes per clique, same number of updates per epoch as 100 nodes, i.e. batch-size 10x less per node) with different inter-clique topologies (variation of \autoref{fig:d-cliques-scaling-cifar10-1000} with 1 class/node instead of 2 shards/node).}
\end{figure}

\autoref{fig:d-cliques-scaling-mnist-1000-1-class-per-node} and \autoref{fig:d-cliques-scaling-cifar10-1000-1-class-per-node} are consistent with \autoref{fig:d-cliques-scaling-mnist-1000} and \autoref{fig:d-cliques-scaling-cifar10-1000}. The less extreme skew enables a slightly faster convergence rate in the case of CIFAR10 (\autoref{fig:d-cliques-scaling-cifar10-1000}).


\subsubsection{Full Intra-Clique Connectivity is Necessary}



\begin{figure}[htbp]
     \centering
\begin{subfigure}[htbp]{0.23\textwidth}
     \centering   
         \includegraphics[width=\textwidth]{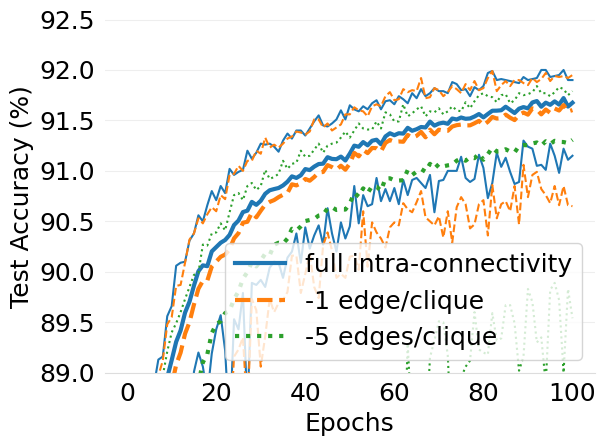}     
\caption{\label{fig:d-cliques-ideal-wo-clique-avg-impact-of-edge-removal} Without Clique Averaging }
\end{subfigure}
\hfill
\begin{subfigure}[htbp]{0.23\textwidth}
     \centering
         \includegraphics[width=\textwidth]{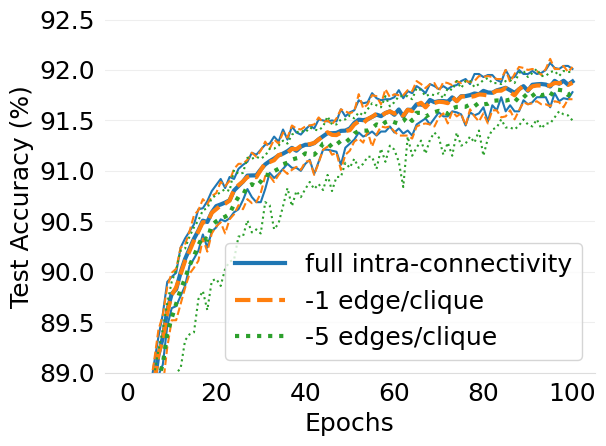}
\caption{\label{fig:d-cliques-ideal-w-clique-avg-impact-of-edge-removal} With Clique Averaging}
\end{subfigure}
\caption{\label{fig:d-cliques-ideal-mnist-intra-connectivity} MNIST: Impact of intra-clique edge removal on D-Cliques (fully-connected) with 10 cliques of 10 heterogeneous nodes (100 nodes) (variation of \autoref{fig:d-cliques-mnist-intra-connectivity} with 1 class/node instead of 2 shards/node). Y axis starts at 89.}
\end{figure}

\begin{figure}[t]
     \centering
\begin{subfigure}[htbp]{0.23\textwidth}
     \centering   
         \includegraphics[width=\textwidth]{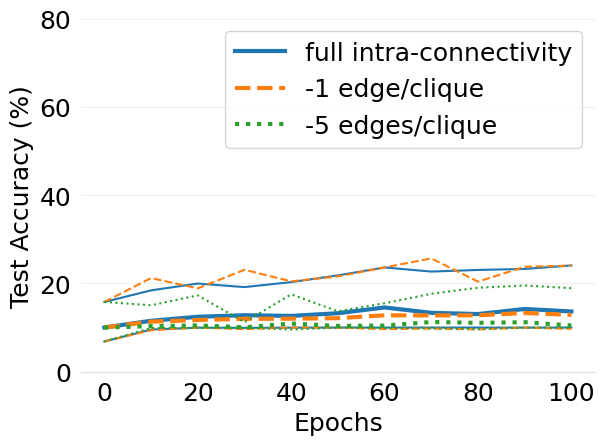}     
\caption{\label{fig:d-cliques-ideal-cifar10-wo-clique-avg-impact-of-edge-removal} Without Clique Averaging }
\end{subfigure}
\hfill
\begin{subfigure}[htbp]{0.23\textwidth}
     \centering
         \includegraphics[width=\textwidth]{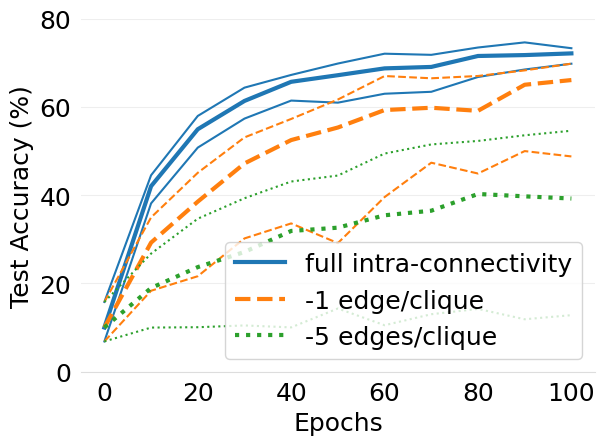}
\caption{\label{fig:d-cliques-ideal-cifar10-w-clique-avg-impact-of-edge-removal} With Clique Averaging}
\end{subfigure}
\caption{\label{fig:d-cliques-ideal-cifar10-intra-connectivity} CIFAR10: Impact of intra-clique edge removal (with momentum) on D-Cliques (fully-connected) with 10 cliques of 10 heterogeneous nodes (100 nodes) (variation of \autoref{fig:d-cliques-cifar10-intra-connectivity} with 1 class/node instead of 2 shards/node).}
\end{figure}

\autoref{fig:d-cliques-ideal-mnist-intra-connectivity} and \autoref{fig:d-cliques-ideal-cifar10-intra-connectivity} show higher variance than \autoref{fig:d-cliques-mnist-intra-connectivity} and \autoref{fig:d-cliques-cifar10-intra-connectivity}, with a significantly lower convergence speed in the case of CIFAR10 (\autoref{fig:d-cliques-ideal-cifar10-intra-connectivity}).


\end{document}